\documentclass[10pt,twocolumn,letterpaper]{article}

\usepackage{cvpr}              
\usepackage{amsmath}
\usepackage{makecell}
\usepackage{tabularx}
\usepackage{cuted}
\usepackage{capt-of}
\usepackage{graphicx}
\usepackage{subcaption}
\usepackage{adjustbox}
\usepackage[accsupp]{axessibility} 
\usepackage{tikz}

\newcommand{\leaveout}[1]{}

\definecolor{cvprblue}{rgb}{0.21,0.49,0.74}
\definecolor{mengli_wo_gpt}{RGB}{191, 64, 191} 
\definecolor{note}{RGB}{0, 255, 255} 
\usepackage[pagebackref,breaklinks,colorlinks,citecolor=cvprblue]{hyperref}

\def\paperID{23} 
\def\confName{CVPR}
\def\confYear{2024}

\title{ExtraNeRF: Visibility-Aware View Extrapolation of \\
Neural Radiance Fields with Diffusion Models}

\author{Meng-Li Shih$^{1}$ \quad 
Wei-Chiu Ma$^{1,2}$ \quad 
Lorenzo Boyice$^{3}$ \quad 
Aleksander Holynski$^{3,4}$ \quad 
Forrester Cole$^{3}$ \quad \\
Brian Curless$^{1,3}$ \quad
Janne Kontkanen$^{3}$\\
$^{1}$ University of Washington \quad $^{2}$ Cornell University \quad $^{3}$ Google Research \quad $^{4}$ UC Berkeley\\
}

\begin{document}
\newcommand{\bbR}{{\mathbb{R}}}
\newcommand{\bK}{\mathbf{K}}
\newcommand{\bX}{\mathbf{X}}
\newcommand{\bY}{\mathbf{Y}}
\newcommand{\bk}{\mathbf{k}}
\newcommand{\bx}{\mathbf{x}}
\newcommand{\by}{\mathbf{y}}
\newcommand{\bhy}{\hat{\mathbf{y}}}
\newcommand{\bty}{\tilde{\mathbf{y}}}
\newcommand{\bG}{\mathbf{G}}
\newcommand{\bI}{\mathbf{I}}
\newcommand{\bg}{\mathbf{g}}
\newcommand{\bS}{\mathbf{S}}
\newcommand{\bs}{\mathbf{s}}
\newcommand{\bM}{\mathbf{M}}
\newcommand{\bw}{\mathbf{w}}
\newcommand{\eye}{\mathbf{I}}
\newcommand{\bU}{\mathbf{U}}
\newcommand{\bV}{\mathbf{V}}
\newcommand{\bW}{\mathbf{W}}
\newcommand{\bn}{\mathbf{n}}
\newcommand{\bv}{\mathbf{v}}
\newcommand{\bq}{\mathbf{q}}
\newcommand{\bR}{\mathbf{R}}
\newcommand{\bi}{\mathbf{i}}
\newcommand{\bj}{\mathbf{j}}
\newcommand{\bp}{\mathbf{p}}
\newcommand{\bt}{\mathbf{t}}
\newcommand{\bJ}{\mathbf{J}}
\newcommand{\bu}{\mathbf{u}}
\newcommand{\bB}{\mathbf{B}}
\newcommand{\bD}{\mathbf{D}}
\newcommand{\bz}{\mathbf{z}}
\newcommand{\bP}{\mathbf{P}}
\newcommand{\bC}{\mathbf{C}}
\newcommand{\bA}{\mathbf{A}}
\newcommand{\bZ}{\mathbf{Z}}
\newcommand{\bff}{\mathbf{f}}
\newcommand{\bF}{\mathbf{F}}
\newcommand{\bo}{\mathbf{o}}
\newcommand{\bc}{\mathbf{c}}
\newcommand{\bT}{\mathbf{T}}
\newcommand{\bQ}{\mathbf{Q}}
\newcommand{\bL}{\mathbf{L}}
\newcommand{\bl}{\mathbf{l}}
\newcommand{\ba}{\mathbf{a}}
\newcommand{\bE}{\mathbf{E}}
\newcommand{\bH}{\mathbf{H}}
\newcommand{\bd}{\mathbf{d}}
\newcommand{\br}{\mathbf{r}}
\newcommand{\bb}{\mathbf{b}}
\newcommand{\bh}{\mathbf{h}}

\newcommand{\btheta}{\bm{\theta}}
\newcommand{\bhh}{\hat{\mathbf{h}}}
\newcommand{\ci}{{\cal I}}
\newcommand{\ct}{{\cal T}}
\newcommand{\co}{{\cal O}}
\newcommand{\ck}{{\cal K}}
\newcommand{\cu}{{\cal U}}
\newcommand{\cS}{{\cal S}}
\newcommand{\cQ}{{\cal Q}}
\newcommand{\cT}{{\cal S}}
\newcommand{\cC}{{\cal C}}
\newcommand{\cE}{{\cal E}}
\newcommand{\cF}{{\cal F}}
\newcommand{\cL}{{\cal L}}
\newcommand{\X}{{\cal{X}}}
\newcommand{\Y}{{\cal Y}}
\newcommand{\cH}{{\cal H}}
\newcommand{\cP}{{\cal P}}
\newcommand{\cN}{{\cal N}}
\newcommand{\cU}{{\cal U}}
\newcommand{\cV}{{\cal V}}
\newcommand{\cX}{{\cal X}}
\newcommand{\cY}{{\cal Y}}
\newcommand{\graph}{{\cal H}}
\newcommand{\bayes}{{\cal B}}
\newcommand{\cx}{{\cal X}}
\newcommand{\cg}{{\cal G}}
\newcommand{\cm}{{\cal M}}
\newcommand{\cM}{{\cal M}}
\newcommand{\cG}{{\cal G}}
\newcommand{\cR}{\cal{R}}
\newcommand{\R}{\cal{R}}
\newcommand{\eig}{\mathrm{eig}}

\newcommand{\bbS}{\mathbb{S}}

\newcommand{\D}{{\cal D}}
\newcommand{\bfp}{{\bf p}}
\newcommand{\bfd}{{\bf d}}

\newcommand{\cv}{{\cal V}}
\newcommand{\ce}{{\cal E}}
\newcommand{\cy}{{\cal Y}}
\newcommand{\cz}{{\cal Z}}
\newcommand{\cb}{{\cal B}}
\newcommand{\cq}{{\cal Q}}
\newcommand{\cd}{{\cal D}}
\newcommand{\bcf}{{\cal F}}
\newcommand{\cI}{\mathcal{I}}

\newcommand{\ut}{^{(t)}}
\newcommand{\up}{^{(t-1)}}

\newcommand{\bpi}{\boldsymbol{\pi}}
\newcommand{\bphi}{\boldsymbol{\phi}}
\newcommand{\bPhi}{\boldsymbol{\Phi}}
\newcommand{\bmu}{\boldsymbol{\mu}}
\newcommand{\bsigma}{\boldsymbol{\sigma}}
\newcommand{\bSigma}{\boldsymbol{\Sigma}}
\newcommand{\bGamma}{\boldsymbol{\Gamma}}
\newcommand{\bbeta}{\boldsymbol{\beta}}
\newcommand{\bomega}{\boldsymbol{\omega}}
\newcommand{\blambda}{\boldsymbol{\lambda}}
\newcommand{\bkappa}{\boldsymbol{\kappa}}
\newcommand{\btau}{\boldsymbol{\tau}}
\newcommand{\balpha}{\boldsymbol{\alpha}}
\def\bgamma{\boldsymbol\gamma}

\newcommand{\prox}{{\mathrm{prox}}}

\twocolumn[{%
\renewcommand\twocolumn[1][]{#1}%
\maketitle
\vspace{-8mm}
\newlength\fta
\setlength\fta{4.2cm}
\begin{center}
    \centering
    \parbox[t]{\fta}{\vspace{0mm}\centering%
      \includegraphics[width=\fta,trim=0 0 0 0,clip]{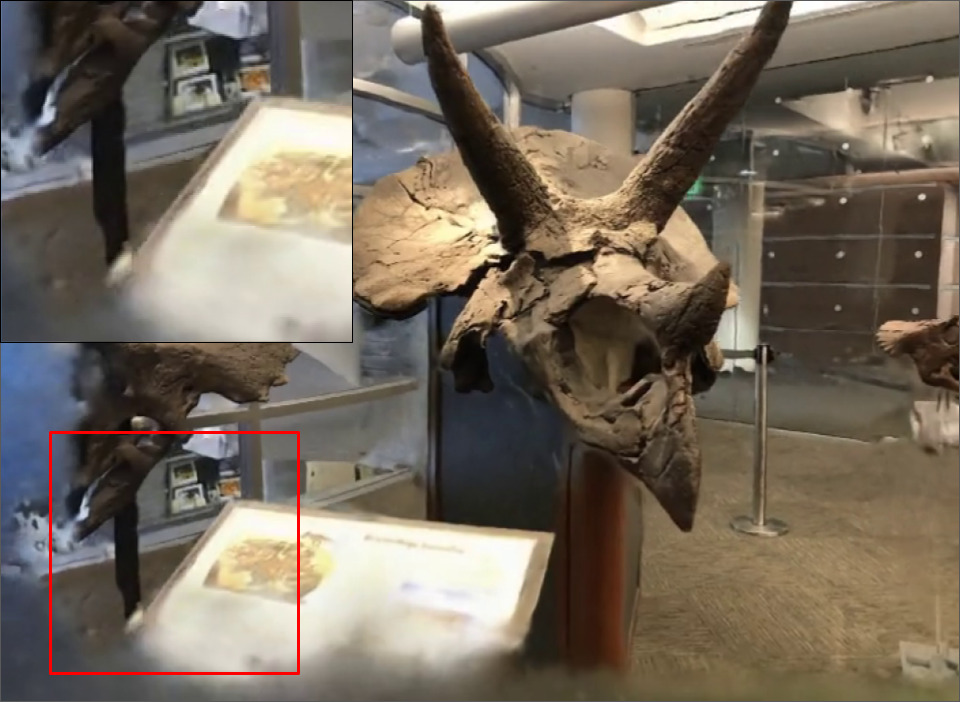}\vspace{-0.5mm}\\%
    	{\small BaseNeRF (View 1)}}%
    \hfill%
    \parbox[t]{\fta}{\vspace{0mm}\centering%
      \includegraphics[width=\fta,trim=0 0 0 0,clip]{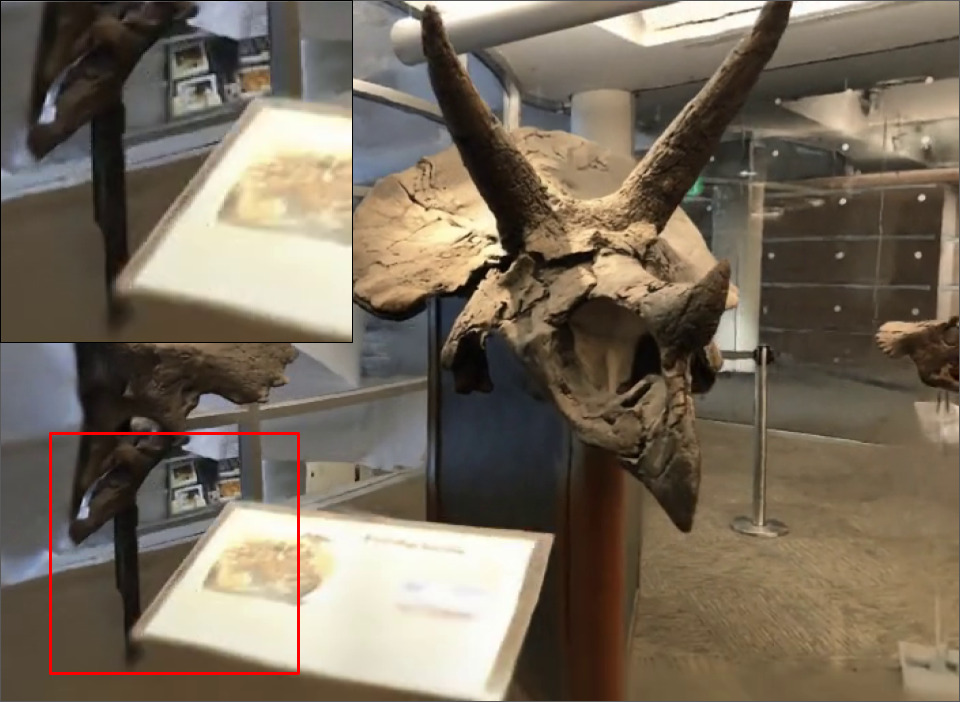}\vspace{-0.5mm}\\%
    	{\small ExtraNeRF (View 1)}}%
    \hfill%
    \parbox[t]{\fta}{\vspace{0mm}\centering%
      \includegraphics[width=\fta,trim=0 0 0 0,clip]{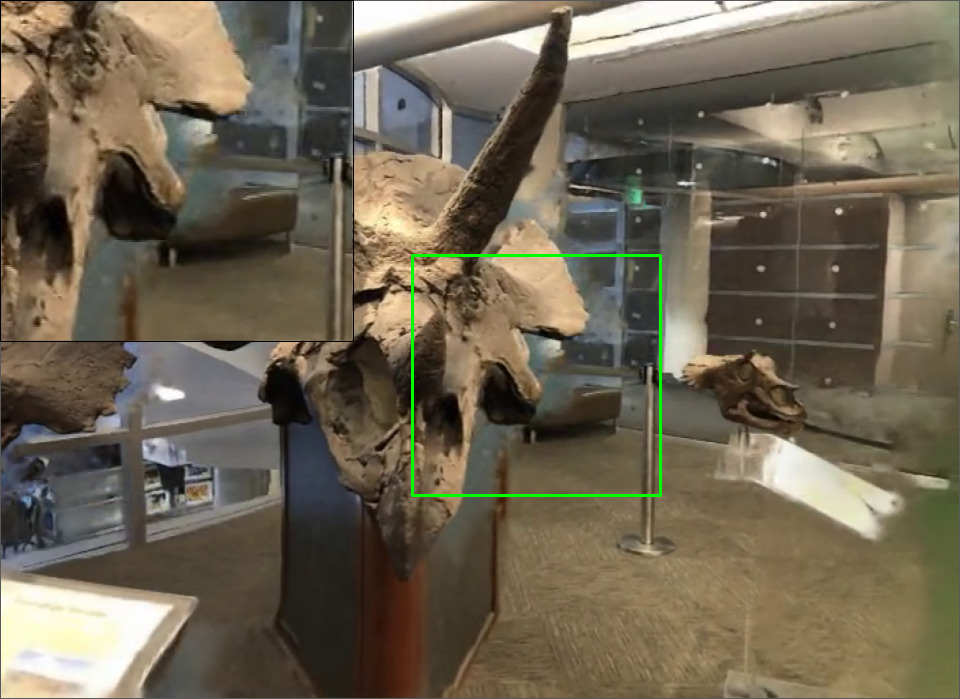}\vspace{-0.5mm}\\%
    	{\small BaseNeRF (View 2)}}%
    \hfill%
    \parbox[t]{\fta}{\vspace{0mm}\centering%
      \includegraphics[width=\fta,trim=0 0 0 0,clip]{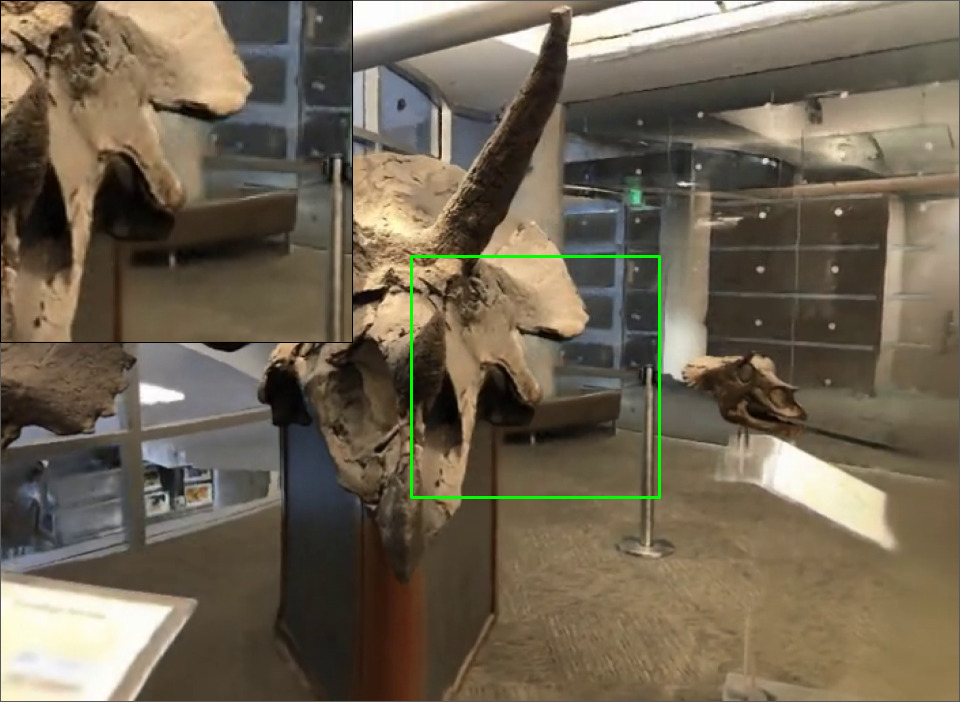}\vspace{-0.5mm}\\%
    	{\small ExtraNeRF (View 2)}}%
        
    \captionof{figure}{
    \textbf{BaseNeRF vs ExtraNeRF:} {We train a BaseNeRF model and our ExtraNeRF model on six input views and render the scene from extrapolated viewpoints. Using our visibility-aware, diffusion-guided inpainting and enhancement modules, we are able to synthesize sharp content in disoccluded regions, whereas the BaseNeRF suffers from blurry results (see the red boxes, green boxes, and the close-up insets)}.  
    }
\label{fig:teaser}
\end{center}



}]
\begin{abstract}
\vspace{-2mm}
We propose ExtraNeRF, a novel method for extrapolating the range of views handled by a Neural Radiance Field (NeRF).  Our main idea is to leverage NeRFs to model scene-specific, fine-grained details, while capitalizing on diffusion models to extrapolate beyond our observed data. A key ingredient is to track visibility to determine what portions of the scene have not been observed, and focus on reconstructing those regions consistently with diffusion models.  Our primary contributions include a visibility-aware diffusion-based inpainting module that is fine-tuned on the input imagery, yielding an initial NeRF with moderate quality (often blurry) inpainted regions, followed by a second diffusion model trained on the input imagery to consistently enhance, notably sharpen, the inpainted imagery from the first pass.  We demonstrate high-quality results, extrapolating beyond a small number of (typically six or fewer) input views, effectively outpainting the NeRF as well as inpainting newly disoccluded regions inside the original viewing volume.  We compare with related work both quantitatively and qualitatively and show significant gains over prior art.
Project page: \url{https://shihmengli.github.io/extranerf-website}
\vspace{-6mm}
\end{abstract}
\section{Introduction}
\label{sec:intro}

Reconstructing a scene from photographs is an important and long-standing problem in computer vision. Recent advances, following the introduction of Neural Radiance Fields (NeRF)~\cite{mildenhall2020nerf} have led to an explosion of progress. Nevertheless, a limitation of NeRF in its base form is that it is far better at interpolating than extrapolating, and requires dense views for the interpolation.  But what if you want to take just a few views, a practical constraint in a live capture setting, and extrapolate beyond them to enable a bit more freedom in viewing the scene?  While there has been significant progress in scene-level sparse NeRF reconstruction, the progress on NeRF-based view extrapolation is primarily limited to object-centric scenarios.  Advances in generative techniques, particularly diffusion models, have demonstrated unforeseen capabilities to synthesize previously unseen imagery. This presents an opportunity to expand the operating range of NeRF more broadly to view extrapolation.

Our core strategy employs neural radiance fields (NeRF~\cite{mildenhall2020nerf}) to capture scene-specific, fine-grained details and utilizes 2D diffusion models~\cite{rombach2022high} to extend the scene beyond the limits of observed data. A straightforward fusion of these technologies initially results in NeRF-rendered images that appear blurry and detail-deficient. This is primarily due to the discord between 2D diffusion priors when applied to a 3D scene from varying perspectives, particularly evident in scene-level view extrapolation where intricate details (such as leaves and branches) are significantly diminished.

To address these challenges, we develope a multi-stage process (see Fig.~\ref{fig:pipeline}) that includes: (1) employing a specialized visibility module to identify all 3D content which is visible from the observed data; (2) utilizing a visibility-aware inpainting module, which is tailored for each scene, to imagine and add plausible 3D content into NeRF for view extrapolation and ensure the content from observed data remains unaltered; and (3) enriching view-consistent details in hallucinated content using a carefully designed diffusion enhancement model.


Through this novel pipeline, we demonstrate high quality view extrapolation from a small number of input views, filling in the newly revealed areas outside the original view volume (see Fig.~\ref{fig:teaser}). Our qualitative and quantitative evaluation show significant gains over previous work.

\section{Related Work}
\label{sec:related_work}

\paragraph{View synthesis:}
Given a set of posed images, the goal of view synthesis is to simulate how a scene would look like from novel viewpoints \cite{chen1993view,levoy1996light,szeliski1998stereo}.
The problem is traditionally formulated as an image-based rendering task \cite{gortler1996lumigraph,zitnick2004high}, and impressive results can be achieved by blending pixel colors across views based on depth maps \cite{chaurasia2013depth} or by compositing images using proxy geometry \cite{kopf2014first}.
Recently, with the help of deep neural networks \cite{zhou2018stereo,tulsiani2018layer,lombardi2019neural,riegler2020free,Riegler2021SVS}, the results have been further improved. 
Together with carefully curated scene representations \cite{,tucker2020single,hu2021worldsheet,wiles2020synsin,shen2022sgam}, researchers have been able to synthesize novel views even from a single image \cite{rockwell2021pixelsynth,watson2022novel,tseng2023consistent}.
Similar to the these recent efforts, our work seeks to extrapolate beyond what is visible and predict the content that is occluded in all images. 
However, instead of relying on deep nets to learn the geometric relationships and hallucinate the content in a purely data-driven fashion, we bake the 3D inductive biases (\eg, visibility) into the pipeline to ground the generation process.
This allows us to generate high-quality, realistic and coherent scene content.

\paragraph{Neural radiance fields (NeRF):}
NeRF \cite{mildenhall2020nerf} has revolutionized the field with its simplicity and extraordinary performance \cite{boss2021nerd,srinivasan2021nerv,zhang2021nerfactor,zhang2022modeling,li2021neural,gao2021dynamic,xian2021space,li2023dynibar}. 
However, existing NeRF-based models tend to be under-constrained, leading to the following limitations: first, they require \emph{dense} observations of the scene; and second, their performance degrades significantly when extrapolating rather than interpolating.
To alleviate these issues, researchers have proposed regularizing the underlying scene representation by data-driven statistics \cite{jain2021putting,niemeyer2022regnerf,wynn2023diffusionerf} or geometry constraints \cite{sparf2023}. 
While these approaches greatly reduce the required number of input images, they still assume that the input views have a wide coverage of the scene. 
The task thus still falls under the view interpolation setup.
In this paper, we focus on a common yet extremely challenging setup in live capture setting where we only have access to a few images with small baselines. 
We show that by carefully integrating generative models with NeRF, we can effectively expand the operating range of NeRF and produce high-quality renderings.
\paragraph{Diffusion models: } Diffusion models \cite{sohl2015deep,ho2020denoising,song2020score,rombach2022high} have drawn wide attention across the vision community due to their capacity and scalability. 
They have demonstrated remarkable performance on a plethora of 2D tasks such as image inpainting \cite{saharia2022palette,lugmayr2022repaint}, deblurring \cite{lee2022progressive,whang2022deblurring}, and have enabled high-quality, diverse image generation \cite{saharia2022photorealistic,rombach2022high}.
By combining with neural rendering \cite{mildenhall2020nerf}, the learned diffusion priors can be further lifted to 3D to enable applications such as text-to-3D \cite{poole2022dreamfusion,lin2023magic3d,wang2023prolificdreamer,sun2023dreamcraft3d} or single-/multi-image 3D generation \cite{tang2023make,qian2023magic123,long2023wonder3d,liu2023one,shi2023zero123++,shi2023mvdream,sargent2023zeronvs}.
Similar to these works, we also leverage diffusion models to synthesize novel views and fuse the generation results back to 3D. 
However, rather than focusing on object-centric setup, we study how to model the 3D content of the scene. 
Furthermore, we explicitly track the visibility across views, which allows us to produce both realistic and consistent 3D reconstructions.
Concurrently with our work, Sargent \etal \cite{sargent2023zeronvs}  also attempt to extrapolate 3D scenes. While their focus is primarily on generating content beyond the visible image boundaries, our approach predicts both disoccluded regions and areas that are not observed.





\section{Preliminaries}
\label{sec:prelim}
\paragraph{Neural radiance fields:} 
A neural radiance field (NeRF~\cite{mildenhall2020nerf}) is an implicit scene representation. 
At its core lies a continuous function $f_{\theta}: \mathbb{R}^3 \times \mathbb{R}^2 \mapsto \mathbb{R}^{+} \times \mathbb{R}^3$, parameterized by a neural network, that maps a 3D point $\bx \in \mathbb{R}^3$ and a view direction $\bd \in \mathbb{R}^2$ to a volume density $\sigma \in \mathbb{R}^{+}$ and an RGB radiance $\bc \in \mathbb{R}^3$.
A NeRF can be rendered into a 2-d image as follows. For each pixel, we cast a ray $\br(s) = \bo + s\bd$ from the camera center $\bo$ through the pixel center in direction $\bd$, and sample a set of 3D points along the ray and query their radiance and density.
Then we aggregate the samples and obtain the color of the pixel via volume rendering:
\begin{align}
\bC(\br) = \sum_{i = 1}^{N_{\br}} T_i(1 - \exp(-\sigma_i \delta_i))\bc_i.
\label{eq:volumetric-rendering}
\end{align}
Here, $\delta_i = s_{i + 1} - s_i$ is the distance between adjacent samples, and $T_i = \exp(-\sum_{j = 1}^{i - 1} \sigma_j \delta_j)$ represents the accumulated transmittance along the ray till $s_i$. 
Intuitively, one can think of $T_i$ as \emph{visibility}, since it is the probability that the ray travels to $s_i$ \emph{without} hitting any other particle.

The volume rendering operation is generic and can be adapted to render other properties of the scene, such as geometry (\ie depth) or visibility (see Sec. \ref{sec:implementation}).
For instance, by replacing the color radiance $\bc_i$ in Eq. \ref{eq:volumetric-rendering} with distance $s_i$, we can compute the expected termination depth:
\begin{align}
\bD(\br) = \sum_{i = 1}^{N_{\br}} T_i(1 - \exp(-\sigma_i \delta_i))s_i.
\end{align}
The neural radiance field $f_\theta$ is learned on a \emph{per-scene basis}.
Given a \emph{dense} set of images, the parameters $\theta$ can be learned by minimizing the discrepancy between target pixel colors $\bC^{\text{target}}(\br)$ and the colors rendered by corresponding rays $\bC(\br)$, \ie $L^{\text{rgb}} = \sum_{\br} \lVert\bC^{\text{target}}(\br) - \bC(\br)\rVert^2_2$. 
If depth information is available, or can be computed with methods such as multi-view stereo,
 one can additionally adopt geometric supervision: $L^{\text{depth}} = \sum_{\br} \lVert\bD^{\text{target}}(\br) - \bD(\br)\rVert^2_2$. 
As we will show in the later sections, explicitly regularizing the geometry of the underlying 3D scene is critical when only a few input images are available. It can also enable better extrapolation to unseen, disoccluded regions.
\paragraph{Diffusion models:}
Diffusion~\cite{sohl2015deep,ho2020denoising,song2020score,rombach2022high} has emerged as a powerful approach for generative image synthesis. Diffusion models rely on the learned denoising module $\Psi(x_t, t, l)$ that takes a noisy input image $x_t$ and possible extra conditioning signals (\eg, text prompts $l$, timestep $t$), and predicts the noise $\epsilon$. By iteratively predicting the noise and subtracting it from the data, the model gradually converts the original noisy data $x_t$ to a target sample of interest $x$.
To train such a model, various levels of Gaussian noise $\epsilon$ are added to  original clean data points, and the denoiser $\Psi$ is tasked with predicting the noise:
\begin{align}
L = \mathbb{E}_{x, t, \epsilon}\lVert\epsilon_{{\Psi}}(x_t, t, l) - \epsilon\rVert^2_2
\label{eq:denoise}
\end{align}
Since training diffusion models from scratch is often costly and requires large amount of data, researchers typically fine-tune pre-trained models for specific tasks using Eq. \ref{eq:denoise} with a smaller, domain-specific dataset.
This fine-tuning can be achieved either by directly adjusting the weights of the denoiser $\Psi$~\cite{hu2021lora, ruiz2023dreambooth} or by introducing additional parameters such as learnable embeddings~\cite{gal2022image}. 

\section{Method}
\label{sec:method}
\begin{figure*}[t]
\centering
\includegraphics[width=1.0\textwidth]{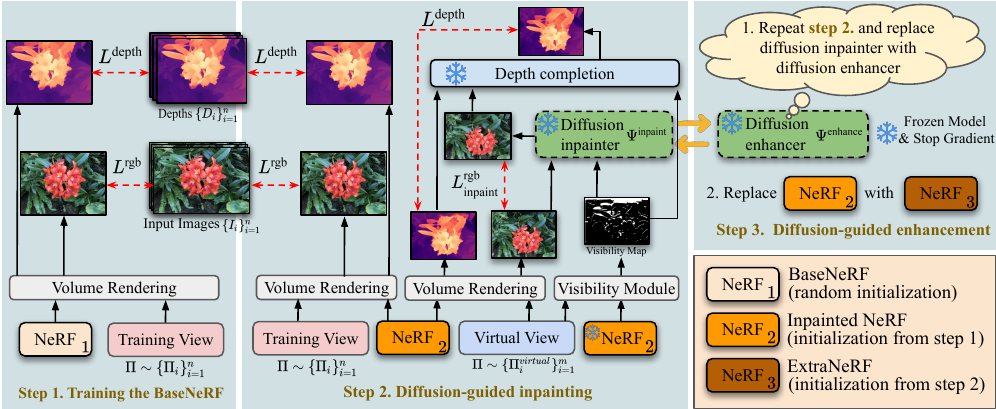}
\caption{
\textbf{Overview of our method:} 
We start from $n$ input images,  their camera poses, and depth maps (predicted as described in Sec.~\ref{sec:method}). In Step 1, we train a BaseNeRF by supervising with this input data. In Step 2, we add supervision from virtual views. We repeatedly inpaint the areas that are unsupervised by the original input views by a diffusion model while continuing to supervise the NeRF with the virtual views. In Step 3, we iterate in similar fashion, but instead of inpainting we apply another diffusion model specifically designed to further improve the detail and color consistency in inpainted regions.}
\label{fig:pipeline}
\end{figure*}

Given a sparse set of images of the scene, our goal is not only to synthesize photo-realistic results between the input views, but also generate high-quality view extrapolations with inpainted disocclusions.


In this section, we first briefly review the basic building blocks of our approach. Next, we explain each component in more detail. Finally, we discuss how we fine-tune our diffusion models and other design choices.

\subsection{Extrapolating Neural Radiance Fields}
We create a NeRF capable of view extrapolation in three steps (see Fig.~\ref{fig:pipeline}): 
\begin{enumerate}
\item {\bf Training the BaseNeRF}: We follow a standard process to train a NeRF on a sparse set of input images.
\item {\bf Diffusion-guided inpainting}:
We iteratively optimize NeRF with virtual views and the original inputs. Each virtual view is rendered from the NeRF and then inpainted using our diffusion model. Then the NeRF can be supervised with this virtual image, backpropagating the newly inpainted regions to the NeRF. Through this iterative process, we construct a consistent neural radiance field that extends beyond the original input images. 
\item {\bf Diffusion guided enhancement}: We find that the previous iterative optimization tends to introduce blur and color drift in the inpainted regions. In the final stage, we use a fine-tuned diffusion model to increase sharpness and improve color consistency in these regions. 
\end{enumerate}

We now describe each component in more detail.

\paragraph{Training the BaseNeRF:}
Given a sparse set of images $\{I_i\}_{i=1}^{n}$ and their associated camera poses $\{\Pi_i\}_{i=1}^{n}$, we first train a BaseNeRF (see Sec. \ref{sec:prelim}). 
Due to the lack of dense multi-view images for effective regularization of the underlying 3D space, we utilize the method proposed in \cite{sparf2023} to compute dense depth maps $\{D_i\}_{i=1}^{n}$ for each input image for geometric supervision. To further reduce ``floater'' artifacts (spuriously reconstructed bits of content in empty regions of the volume), we incorporate distortion loss \cite{barron2021mip} and hash decay loss \cite{barron2023zip} and apply gradient scaling~\cite{philip2023floaters} to regularize the learning procedure.


\paragraph{Diffusion-guided Inpainting: }
Once we have the BaseNeRF, the next step is to augment it such that it can handle extrapolated viewpoints.

To do this, we repeatedly optimize the NeRF over the set of original views and virtual views that extend beyond the original viewing domain. For each virtual view, we render it using the NeRF and then use a diffusion inpainting model $\Psi^{\text{inpaint}}$ to predict the unobserved regions. 

\begin{figure}[t]
  \centering
\includegraphics[width=\linewidth]{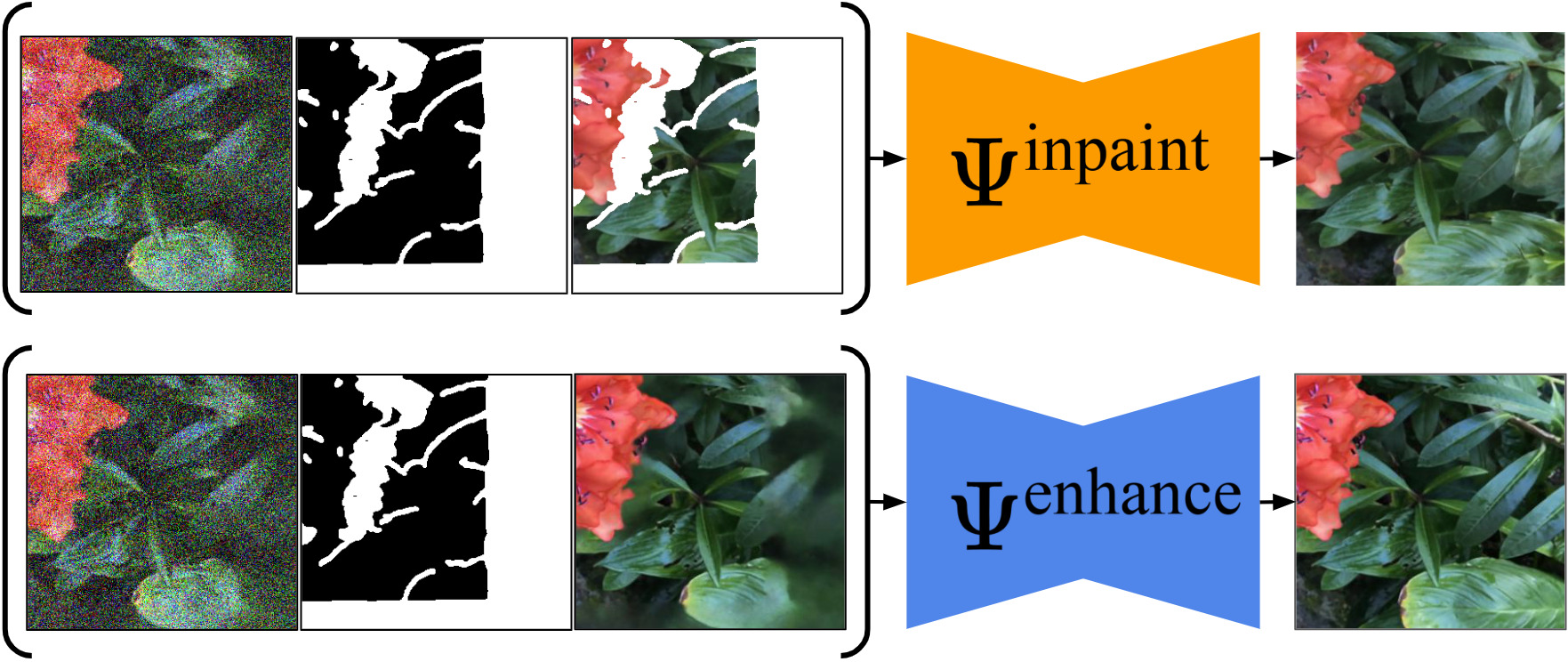}
\caption{The input triplet of diffusion model consists of noisy-image, mask, and an guidance image. While masked pixels of guidance images of ${\Psi}^{\text{inpaint}}$ are erased, they are preserved as the guidance for ${\Psi}^{\text{enhance}}$. 
}
\label{fig:diffusion_io}
\end{figure}

As our inpainting module\ $\Psi^{\text{inpaint}}$, we adopt the inpainting variant of latent diffusion from~\cite{rombach2022high}, which we further fine-tune on a per-scene basis~(see Sec.~\ref{subsec:fine_tuning}).  To limit the inpainting to the unobserved regions (\eg areas where NeRF lacks supervision), our diffusion inpainter\ $\Psi^{\text{inpaint}}$ takes three inputs: noisy image, visibility mask, and masked clean image that lacks data in areas to inpaint (see Fig.~\ref{fig:diffusion_io}).  The visibility masks are computed by checking whether the 3D sample points along the ray at each pixel have been observed in the training images (see Sec. \ref{sec:implementation}). 

For each virtual view, we also inpaint the depth conditioned on the inpainted color image using a depth completion network (see Sec.~\ref{sec:implementation}).   

Once the image and depth for the virtual view are inpainted, they are used to further supervise the NeRF through $L_{\text{inpaint}}^{\text{rgb}}$ and $L^{\text{depth}}$ respectively (see Fig.~\ref{fig:pipeline}). $L_{\text{inpaint}}^{\text{rgb}}$ is computed as follows:
\begin{align}
L_{\text{inapint}}^{\text{rgb}} = \sum_{\br} w(t)|\bC^{\text{inpaint}}(\br) - \bC(\br)|,
\label{eq:inpaint_rgb_loss}
\end{align}
where $w(t)$ is a noise-level dependent weighting function, $\bC^{\text{inpaint}}$ is the inpainted colors and $\bC$ is the rendered image from NeRF.
We chose to run small number of diffusion denoising steps on each virtual view at the time (\eg 10),  but we repeat the whole process by iterating over the views several times. 



Note that while inpainting in multiple views separately could lead to inconsistencies, our iterative approach does converge, because at each virtual view the diffusion process is bootstrapped via the noisy image that is re-estimated from the continuously improving NeRF on every iteration. This is similar to~\cite{poole2022dreamfusion}, although in our work we opted to run more than one step of diffusion before we move to a new view. 

\paragraph{Diffusion-guided enhancement: }

While the iterative inpainting converges into a consistent result, we have observed that some blurriness and color drift may still occur in the NeRF after the inpainting stage. 


To alleviate this, we utilize a diffusion-based enhancement model, $\Psi^{\text{enhance}}$, which has the same architecture as $\Psi^{\text{inpaint}}$ but specifically trained for the enhancement (see Sec.~\ref{subsec:fine_tuning}).

Similar to inpainting, we use an iterative approach to update our NeRF. In each training iteration we 1) render the image and compute the visibility mask from the NeRF, 2) create a triplet of input data from the rendered image and visibility mask, and 3) leverage our $\Psi^{\text{enhance}}$ model to generate an enhanced image from the triplet. In contrast to the inpainting process, we do not mask out the pixels in the intact rendered image (see Fig.~\ref{fig:diffusion_io}). Instead, we want $\Psi^{\text{enhance}}$ to enhance detail in these areas. Once the enhanced image is generated, we then complete the depth. Finally, we supervise the NeRF following steps similar to the inpainting stage but replace $L_{\text{inpaint}}^{\text{rgb}}$ with $L_{\text{enhance}}^{\text{rgb}}$. $L_{\text{enhance}}^{\text{rgb}}$ is almost identical to $L_{\text{inpaint}}^{\text{rgb}}$ except that we replace $\bC^{\text{inpaint}}$ with $\bC^{\text{enhance}}$ (i.e. enhanced colors).
As shown in Fig.~\ref{fig:ablations}, this process can improve detail and overall image quality. 

\begin{figure}[t] 
  \centering
\includegraphics[width=\linewidth,trim=0 0 0 0, clip]{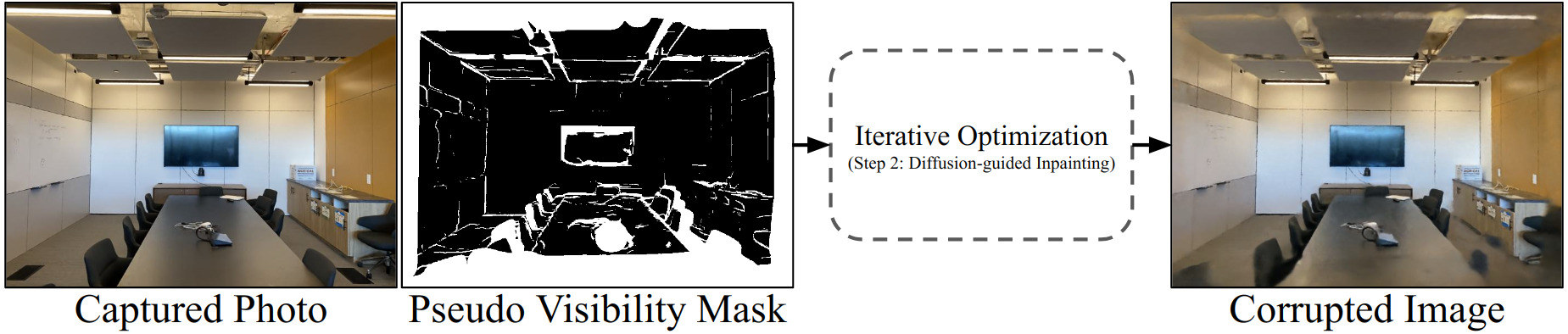}
\caption{Illustration of data collection for enhancement model.
We draw a pseudo visibility mask in a captured photo. Ground-truth supervision in the mask is replaced by inpainting supervision when we iteratively optimize NeRF. The optimization corrupts pixels in the mask when rendered with NeRF. A captured photo along with several corrupted images from different optimization iterations can used to train ${\Psi}^{\text{enhance}}$}
\label{fig:data_aug_enhance}
\end{figure}
\subsection{Fine-tuning the Diffusion Models: }
\label{subsec:fine_tuning}
As mentioned earlier, we use the inpainting-variant of the latent diffusion model~\cite{rombach2022high} for both inpainting and enhancement. For the best quality it is essential to fine-tune both $\Psi^{\text{inpaint}}$ and $\Psi^{\text{enhance}}$ for the scene and their respective tasks using our sparse set of input images  $\{I_i\}_{i=1}^{n}$. This is perhaps obvious in the case of $\Psi^{\text{enhance}}$ as its task is not exactly inpainting, but more similar to deblurring. However, as shown in Fig~\ref{fig:ablations}, scene-specific fine-tuning is also important for the inpainting module.

To fine-tune these two models $\Psi^{\text{inpaint}}$ and $\Psi^{\text{enhance}}$, we devise a process to produce ground truth training data using our input images. The first step is to create visibility masks similar to those that would occur in the virtual views. For each training image, we compute a corresponding pseudo visibility mask by checking if pixels are visible in all other training views. Pixels not visible in one more view are treated as disocclusions, and we fine-tune $\Psi^{\text{inpaint}}$ by asking it to inpaint the regions under these masks. 

We fine-tune both models using standard diffusion loss (Eq. \ref{eq:denoise}) following the DreamBooth~\cite{ruiz2023dreambooth} pipeline.

To produce training data for $\Psi^{\text{enhance}}$ a further step is needed. To produce corrupted input for enhancement, we optimize a NeRF by intentionally replacing supervision from input viewpoints with inpainting supervision from $\Psi^{\text{inpaint}}$ for pixels in the pseudo-disocclusion masks. We find that this adequately simulates the blur and color drift that the $\Psi^{\text{enhance}}$ is tasked to reduce. See Fig.~\ref{fig:data_aug_enhance} for an example.

\subsection{Implementation details}
\label{sec:implementation}

\paragraph{Visibility map:}
The visibility map indicates whether the 3D points corresponding to the pixels of a virtual view 
are visible in the input images. They might be hidden if they are outside the input view frustums or occluded by a closer object.

It plays a critical role in our system as it helps us determine which areas are unobserved in the original images and require inpainting. As indicated in Sec. \ref{sec:prelim}, the accumulated transmittance from NeRF encodes essential visibility information. This enables us to estimate the visibility of any 3D point w.r.t the input views.

To compute the visibility map for a single pixel of a virtual view, we first construct a ray through that pixel. For each sampled 3D point along this ray, we then compute the  transmittance towards each training view (e.g. another ray march). To aggregate the transmittance values across the input views, we simply select the second largest value. This is based on the rationale that the geometry of a 3D point is only reliable if observed by at least two views (the minimum for triangulation). If a 3D point is seen by only one training view, its estimated depth might be unreliable. Finally, these aggregated transmittance samples are aggregated together to the visibility map pixel by volume rendering, similarly to color values.



\begin{figure}[t]
    \captionsetup[subfigure]{labelformat=empty}
    \centering
    \begin{subfigure}[b]{0.155\textwidth}
        \centering
        \includegraphics[width=\textwidth]{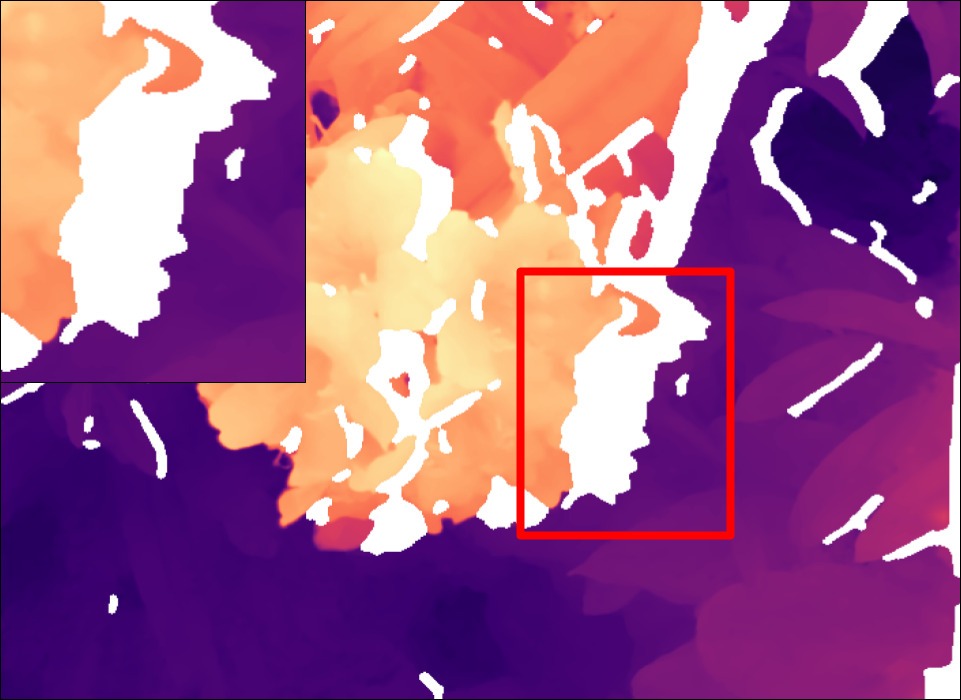}
        \caption{Masked Depth}
    \end{subfigure}\hfill
    \begin{subfigure}[b]{0.155\textwidth}
        \centering
        \includegraphics[width=\textwidth]{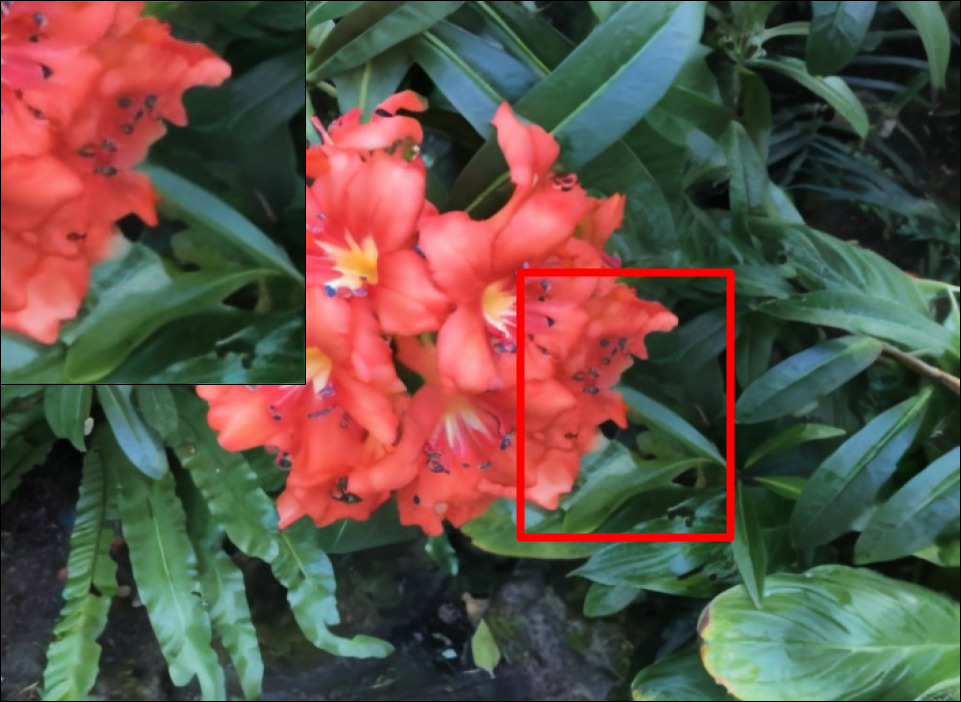}
        \caption{Guidance Image}
    \end{subfigure}\hfill
    \begin{subfigure}[b]{0.155\textwidth}
        \centering
        \includegraphics[width=\textwidth]{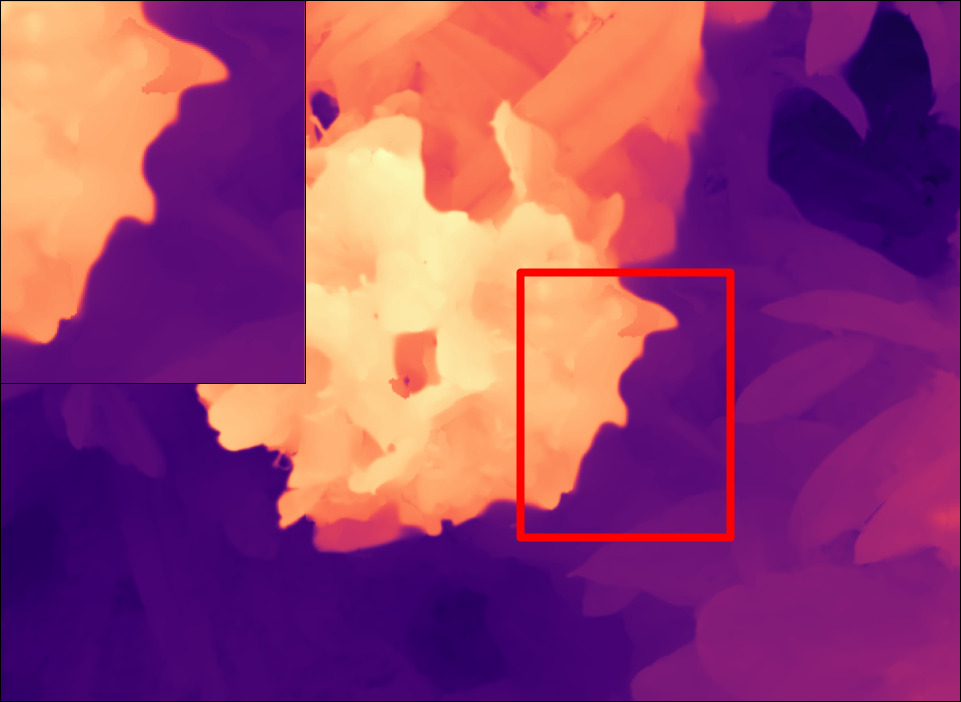}
        \caption{Completed Depth}
    \end{subfigure}\hfill
\caption{The depth completion model takes a masked depth along with a guidance image as input and completes the depth in the masked region using the guidance of the RGB image.}

\label{fig:depth_completion}
\end{figure}
\paragraph{Depth completion module: }
We develop a depth completion module to complete the depth maps for virtual views required by $L^{depth}$ (Fig.~\ref{fig:depth_completion}). The depth completion network takes the inpainted RGB image, visibility mask, and masked depth-map as input, and inpaints depth map in the masked region. The model is based on the pretrained weights of MiDaS-v3~\cite{birkl2023midas} with two additional input channels for the input mask and masked depth-map. The model is fine-tuned with a self-supervised approach on the Places2 dataset~\cite{zhou2016places} (see Suppl. for details). 

\paragraph{Hyper-parameters:}
We fine-tune our inpainting and enhancement models for 500 iterations with a learning rate of 5e-6 for the diffusion U-Net and 4e-5 for the LoRA layer of the text encoder. Our NeRF uses Instant-NGP~\cite{mueller2022instant} as the backbone, with scene contraction~\cite{barron2021mip} to handle unbounded scenes. We propose a 3-stage pipeline to train our NeRF.
In step 1, we train the BaseNeRF for 5000 iterations using $L^{\text{rgb}}$ and $L^{\text{depth}}$ with an initial learning rate of 1e-2, gradually decreasing to 3e-4.
In step 2, in addition to $L^{\text{rgb}}$ and $L^{\text{depth}}$, the NeRF also receives supervision from the inpainting model $\Psi^{\text{inpaint}}$ via $L^{\text{rgb}_{\text{inpaint}}}$ using the virtual views for 500 iterations.
In Stage 3, we supervise the NeRF for another 500 iterations but replace the inpainting model $\Psi^{\text{inpaint}}$ with the enhancement model $\Psi^{\text{enhance}}$.

\paragraph{Time consumption:}
In our experiments, we used one A100 GPU. Step 1, the bottleneck, involved 6 hrs of training Sparf~\cite{sparf2023} with early stopping at 30K iterations to create depth maps, and 8 minutes to train the BaseNeRF model. Steps 2 and 3 typically took 2-3 hrs, including 1 hour for data collection and fine-tuning two diffusion models, as well as 1-2 hrs for training the ExtraNeRF model. Our total optimization time is shorter than that of NeRF~\cite{mildenhall2020nerf}, which can take a day or more when running on a single GPU.

\section{Experiments} \label{sec:experiments}

\subsection{Experimental setup}
\paragraph{LLFF Datasets: }
We primarily utilize the LLFF dataset to demonstrate the effectiveness of our method. This dataset offers two settings for the training/test split that we have explored. In the first protocol, our goal is to assess performance in the task of view extrapolation. Therefore, 6 out of 30-40 images, whose viewpoints are closest to the center position, are chosen as the training set, and 8 images, whose viewpoints are farthest from the center position, are chosen as the test set (see Tab.~\ref{tab:quantitative_llff}). The second protocol follows the standard few-shot view synthesis setup~\cite{niemeyer2022regnerf} (see Tab.~\ref{tab:quantitative_llff_original_protocol_rebutt}).
\paragraph{Tanks\ \&\ Temples Datasets:}
We also utilize the Tanks \& Temples dataset~\cite{Knapitsch2017} to demonstrate our method's capability to manage more complex scenes in real-world settings. The data processed by NeRF++~\cite{zhang2020nerf++} serves as our basis. In each scene, we select 3-5 nearby views as the training set and choose another 5-6 views whose viewpoints surround the training viewpoints, as the test set.


\paragraph{Metrics: }
We adopt the same metrics as~\cite{mirzaei2023reference}, since our goals, akin to theirs, involve evaluating the performance of synthesized 3D content. Accordingly, we utilize two sets of metrics: full-reference (FR) and no-reference (NR). For the FR metrics, we exclusively use LPIPS\cite{zhang2018perceptual}, KID~\cite{binkowski2018demystifying}, and also include PSNR and SSIM~\cite{wang2004image} for a comprehensive assessment. However, it's noteworthy that PSNR and SSIM are not considered reliable metrics for evaluating generative tasks~\cite{sargent2023zeronvs, deng2023nerdi, chan2023genvs}. For NR metrics, MUSIQ~\cite{ke2021musiq} is employed to assess the visual quality of rendered images.
\paragraph{Baselines: }
We compare our method with six related baselines for which code is available:
(1) Sparf~\cite{sparf2023}, one of the state-of-the-art (SOTA) methods for sparse view reconstruction.
(2) FreeNeRF~\cite{Yang2023FreeNeRF}, another SOTA method for sparse view reconstruction.
(3) DiffusioNeRF~\cite{wynn2023diffusionerf}, which employs a patch-wise diffusion model to provide RGB and depth supervision for a NeRF.
(4) SPIn-NeRF~\cite{spinnerf}, aimed at inpainting unobserved content behind an object in 3D, given a complete object mask. In our setting, where no object mask exists, we substitute it with a visibility mask, denoted as *SPIn-NeRF.
(5) *SinNeRF~\cite{xu2022sinnerf}, capable of extrapolating views in 3D from a single image and an accurate depth map. For a fair comparison, we provide RGB supervision from all images in the training set.
(6) *SDS~\cite{poole2022dreamfusion} loss, widely used in 3D content generation. Here, we substitute the color supervision from the inpainted image with SDS loss.

\subsection{LLFF}
\label{subsec:llff_exp}
\paragraph{Comparison of view extrapolation: }
\begin{figure*}[t] 

\captionsetup[subfigure]{labelformat=empty}
\centering
\begin{subfigure}{.245\textwidth}
  \centering
  \includegraphics[width=\linewidth]{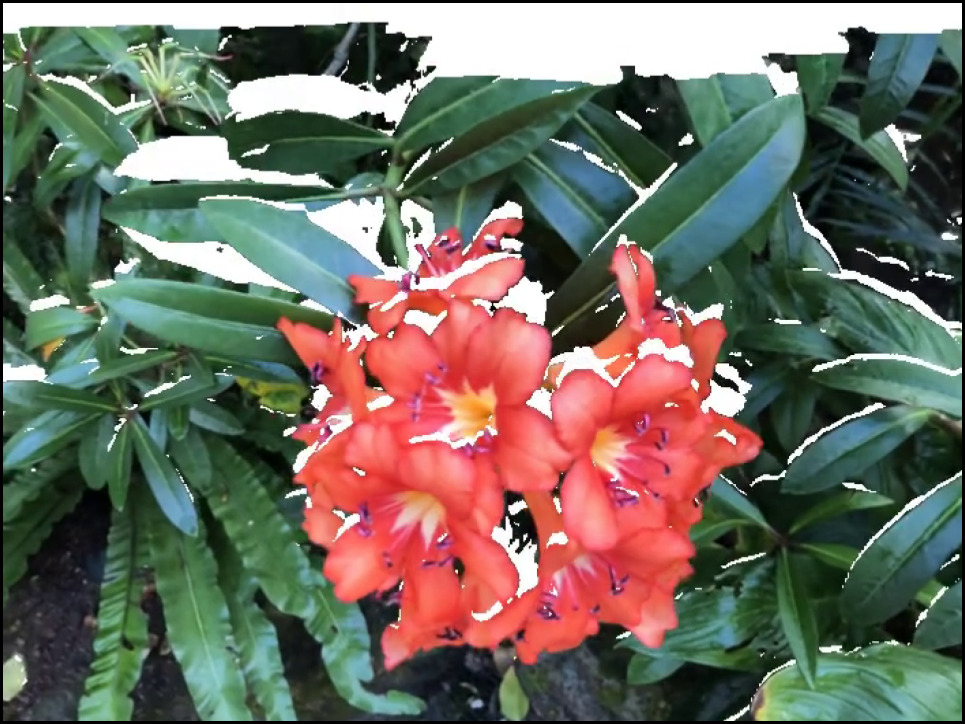}
  \caption{Image \& Visibility mask}
\end{subfigure}\hfill%
\begin{subfigure}{.245\textwidth}
  \centering
  \includegraphics[width=\linewidth]{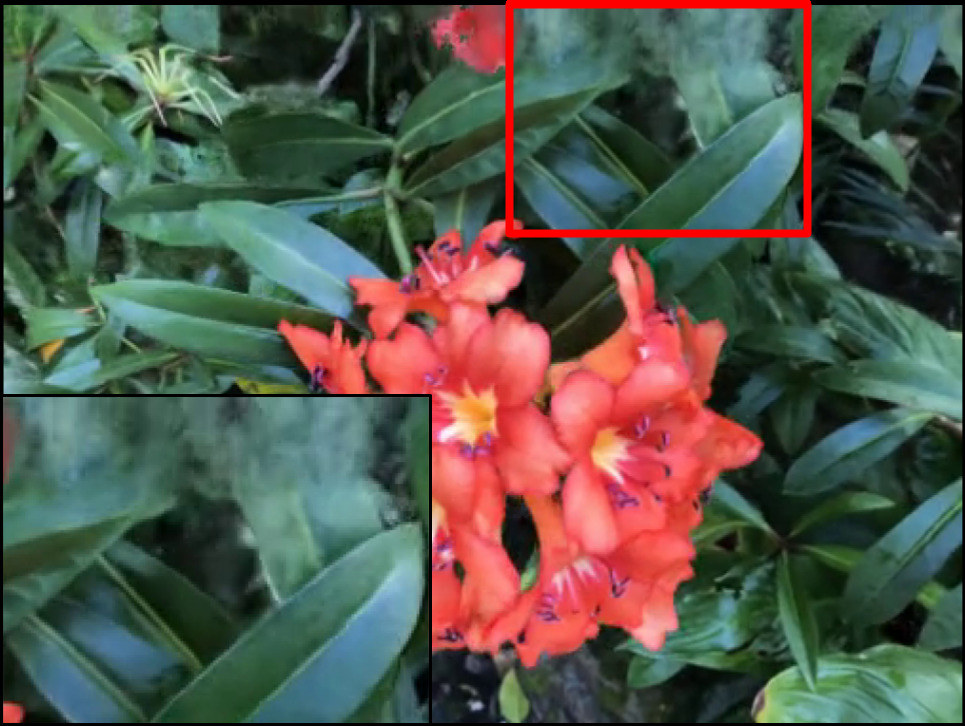}
  \caption{Sparf}
\end{subfigure}\hfill%
\begin{subfigure}{.245\textwidth}
  \centering
  \includegraphics[width=\linewidth]{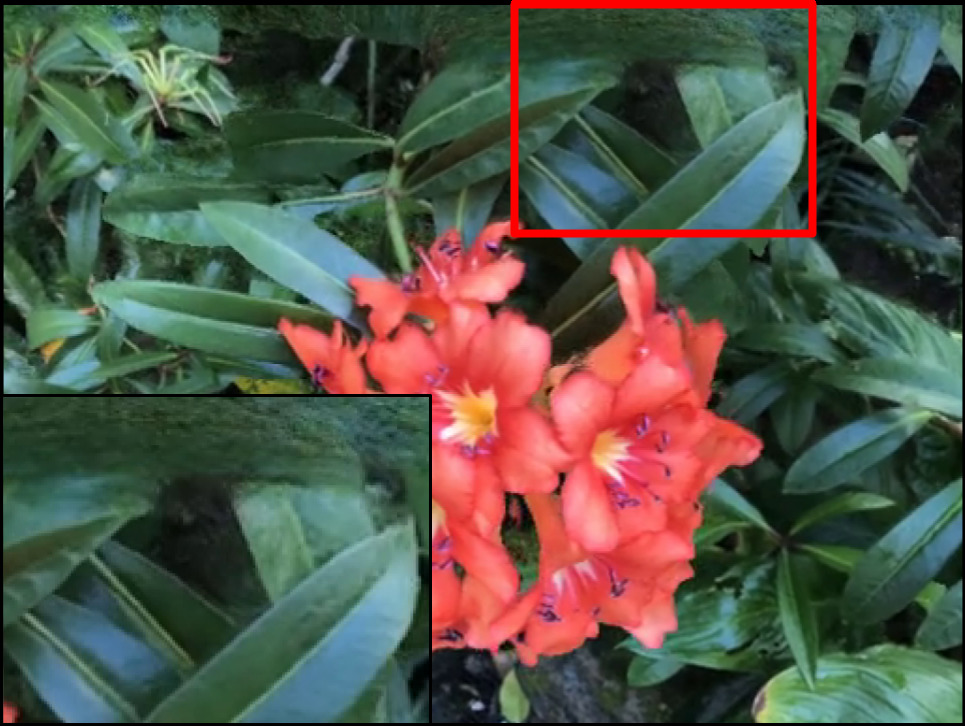}
  \caption{DiffusioNeRF}
\end{subfigure}\hfill%
\begin{subfigure}{.245\textwidth}
  \centering
  \includegraphics[width=\linewidth]{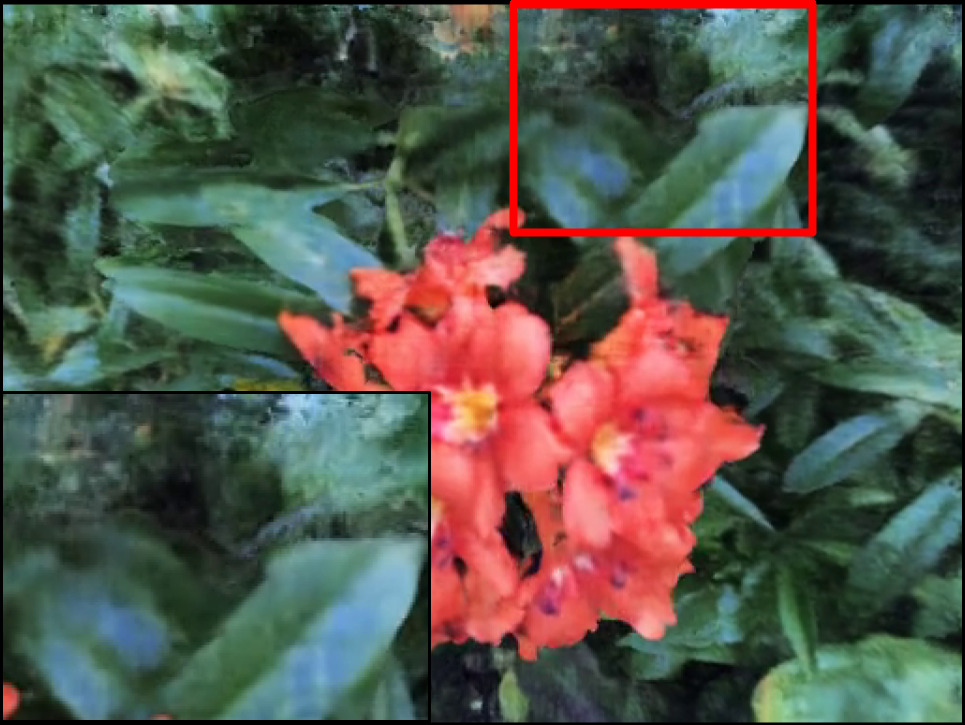}
  \caption{*SinNeRF}
\end{subfigure}\hfill
\begin{subfigure}{.245\textwidth}
  \centering
  \includegraphics[width=\linewidth]{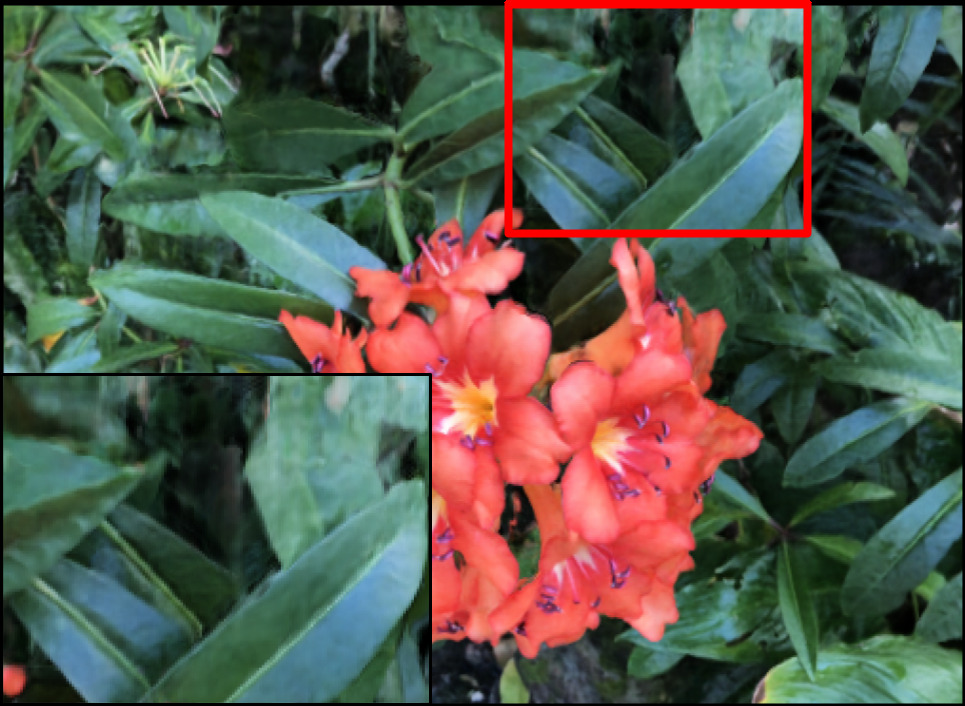}
  \caption{FreeNeRF}
\end{subfigure}\hfill%
\begin{subfigure}{.245\textwidth}
  \centering
  \includegraphics[width=\linewidth]{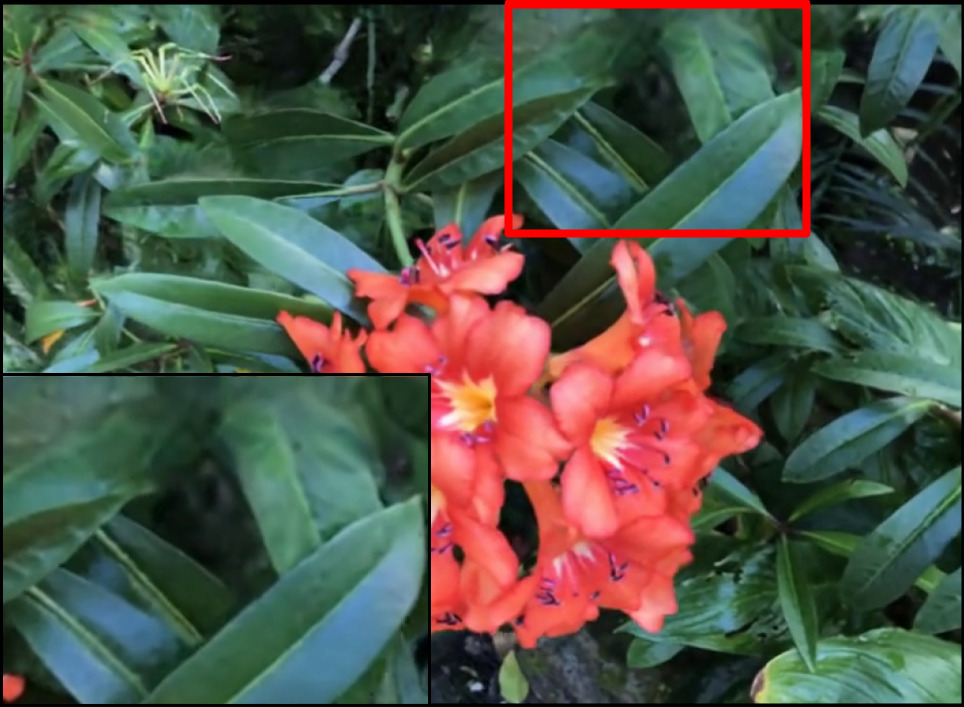}
  \caption{*SPIn-NeRF}
\end{subfigure}\hfill
\begin{subfigure}{.245\textwidth}
  \centering
  \includegraphics[width=\linewidth]{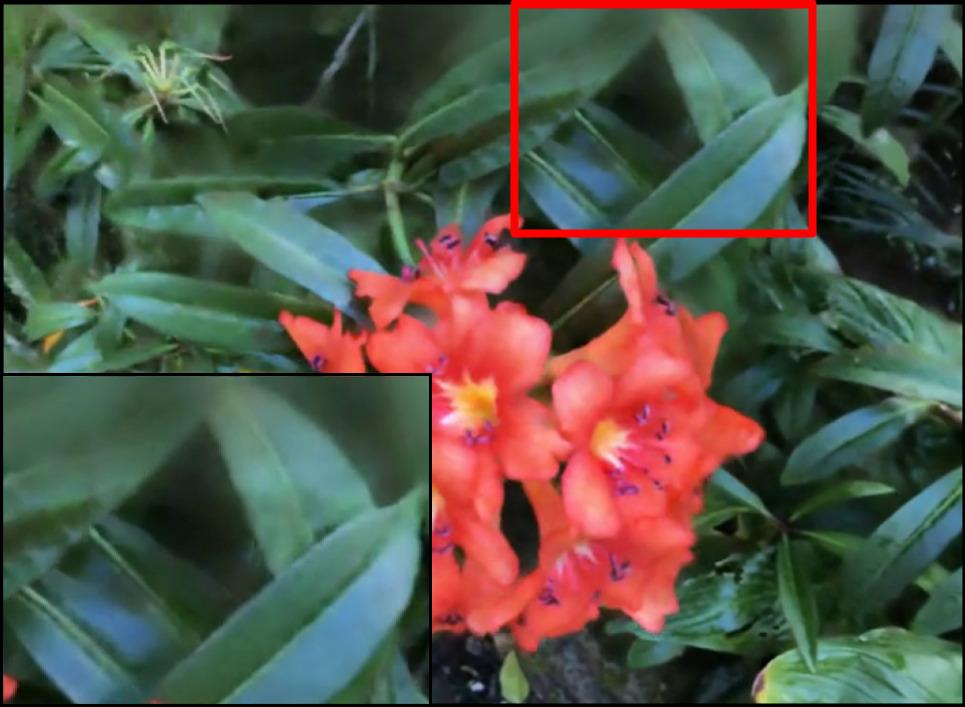}
  \caption{*SDS}
\end{subfigure}\hfill
\begin{subfigure}{.245\textwidth}
  \centering
  \includegraphics[width=\linewidth]{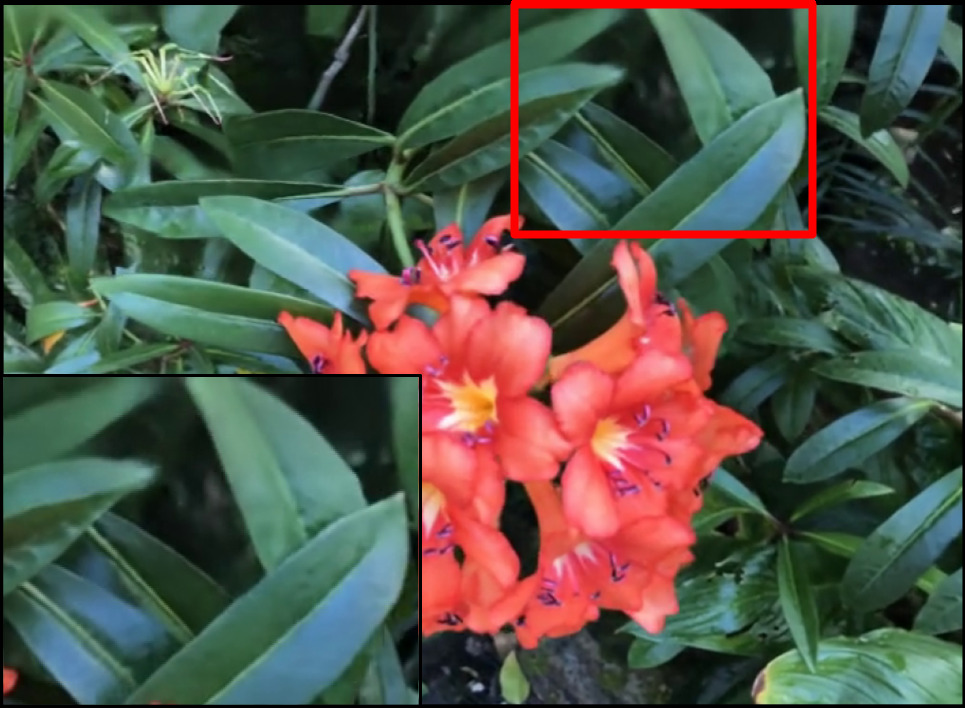}
  \caption{Ours}
\end{subfigure}
\begin{subfigure}{.245\textwidth}
  \centering
  \includegraphics[width=\linewidth]{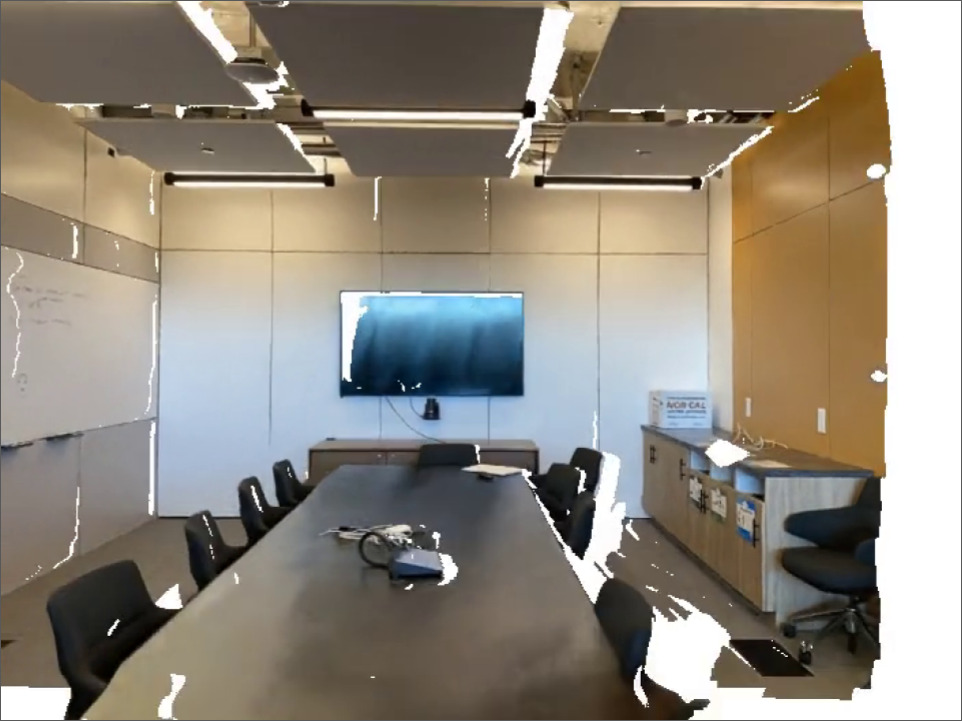}
  \caption{Image \& Visibility mask}
\end{subfigure}\hfill%
\begin{subfigure}{.245\textwidth}
  \centering
  \includegraphics[width=\linewidth]{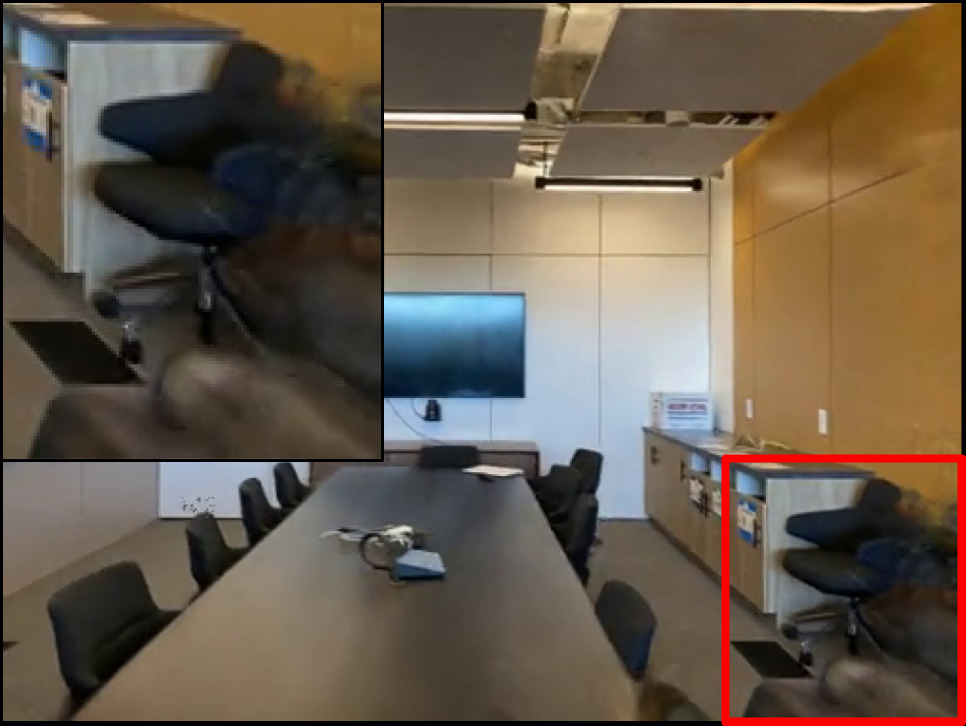}
  \caption{Sparf}
  \label{fig:sfig2}
\end{subfigure}\hfill%
\begin{subfigure}{.245\textwidth}
  \centering
  \includegraphics[width=\linewidth]{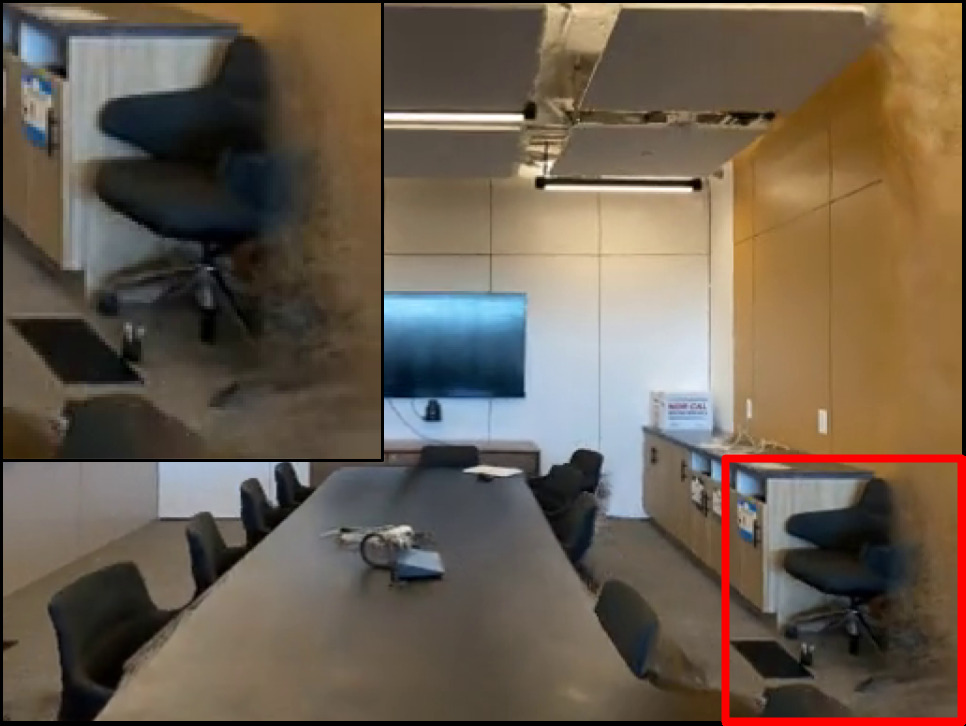}
  \caption{DiffusioNeRF}
\end{subfigure}\hfill%
\begin{subfigure}{.245\textwidth}
  \centering
  \includegraphics[width=\linewidth]{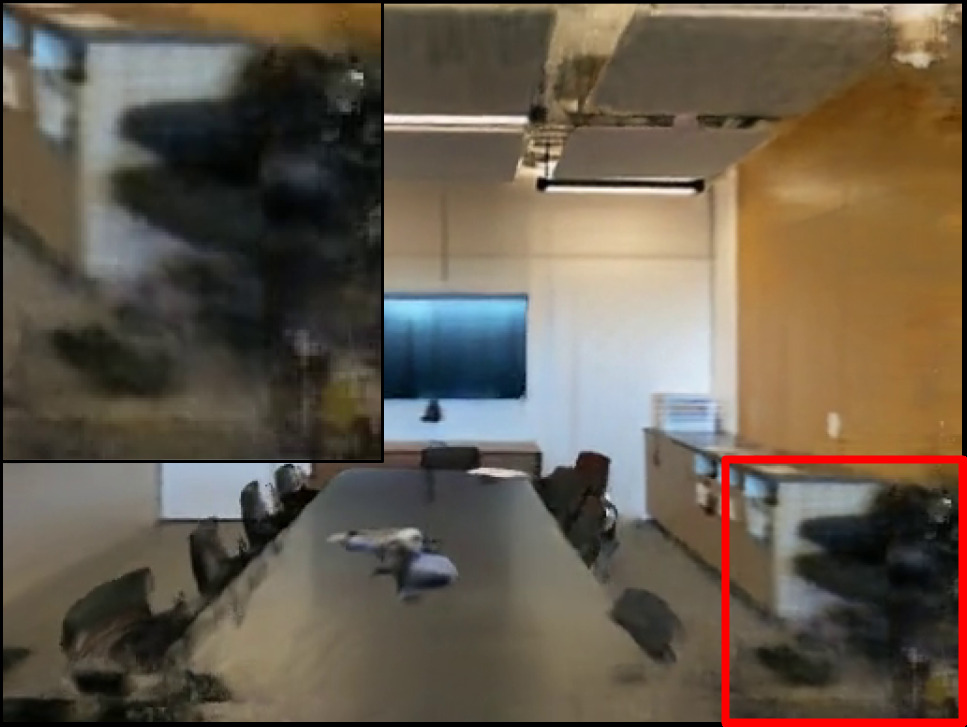}
  \caption{*SinNeRF}
\end{subfigure}\hfill
\begin{subfigure}{.245\textwidth}
  \centering
  \includegraphics[width=\linewidth]{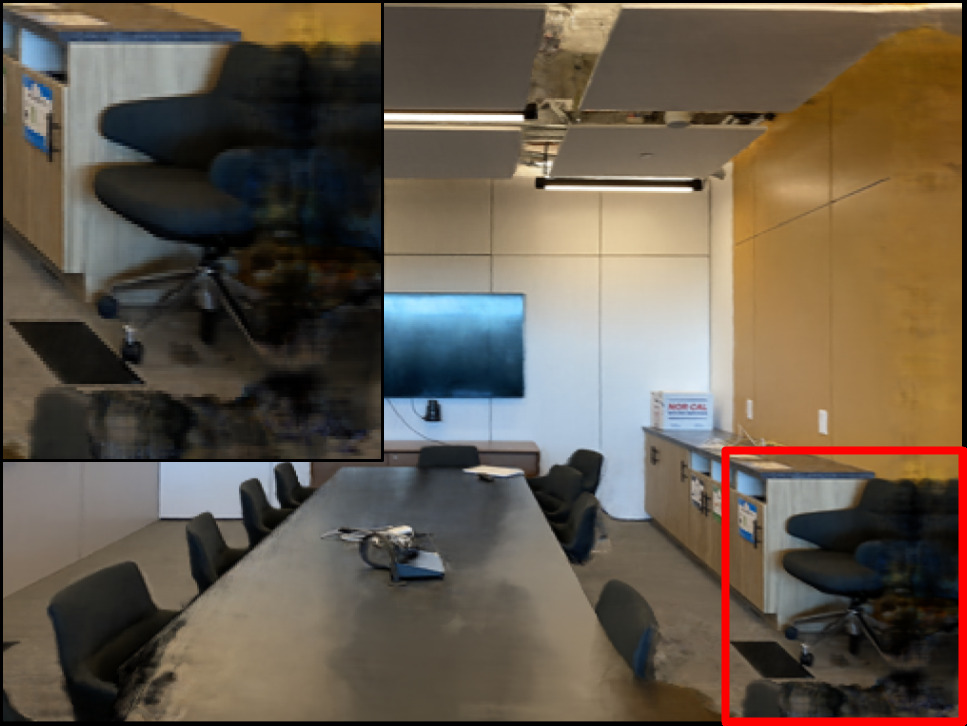}
  \caption{FreeNeRF}
\end{subfigure}\hfill%
\begin{subfigure}{.245\textwidth}
  \centering
  \includegraphics[width=\linewidth]{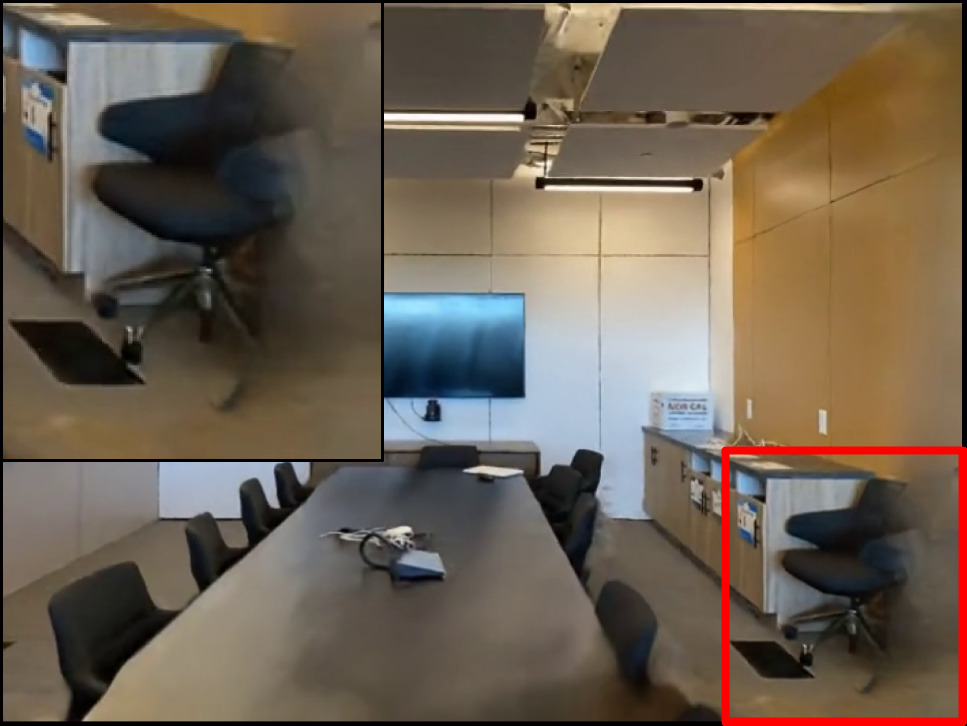}
  \caption{*SPIn-NeRF}
\end{subfigure}\hfill
\begin{subfigure}{.245\textwidth}
  \centering
  \includegraphics[width=\linewidth]{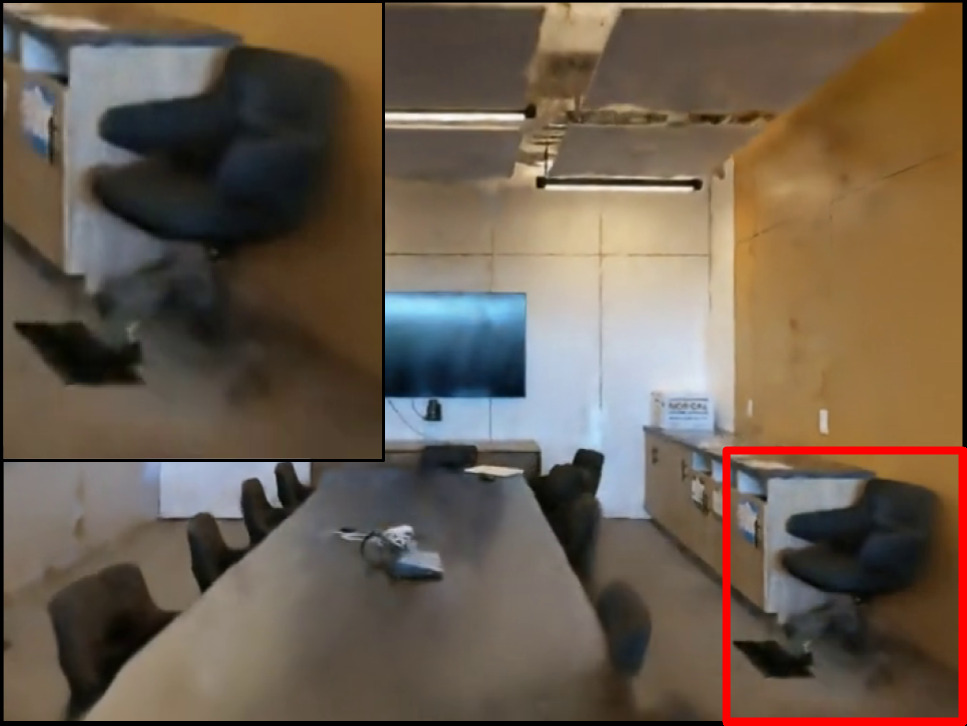}
  \caption{*SDS}
\end{subfigure}\hfill
\begin{subfigure}{.245\textwidth}
  \centering
  \includegraphics[width=\linewidth]{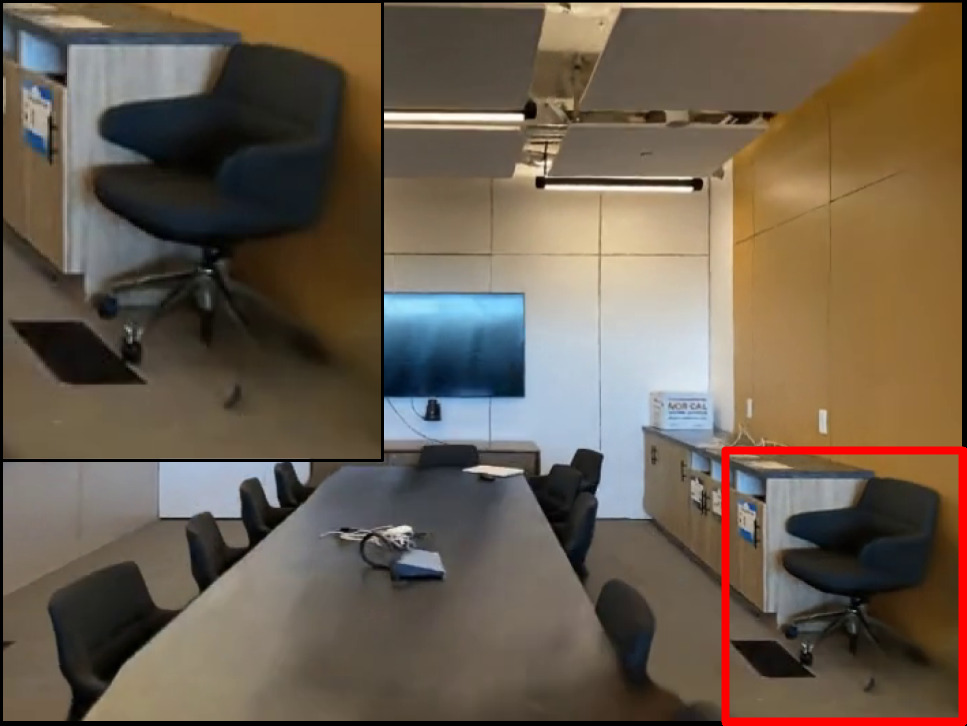}
  \caption{Ours}
\end{subfigure}
\caption{\textbf{Qualitative results of view extrapolation on LLFF dataset.} 
We present the image masked by the visibility mask on the left and the extrapolated results on the right. In comparison to the baselines, our results are significantly sharper and align coherently with the existing scene content. We are able to accurately reconstruct the structure of the leaves (top) and the partially observed chair (bottom), in contrast to the baselines, which yield blurry outcomes and struggle to differentiate between the foreground and background.}
\label{fig:qualitative}

  \label{fig:qualitative}
\end{figure*}
\begin{table}[t]
\caption{Quantitative comparison of view extrapolation.}%
\label{tab:quantitative_llff}
\centering
\small
\setlength{\tabcolsep}{2pt}
\begin{tabular}{l|ccccc}
\toprule
    Methods        &   PSNR $\uparrow$   &   SSIM $\uparrow$    &   LPIPS $\downarrow$    &   KID $\downarrow$ &   MUSIQ $\uparrow$    \\ 
\midrule
    Sparf          &   20.38            &   0.650              &   0.324                 & 0.0199             &  40.32  \\ 
    FreeNeRF       &   20.16           &   0.663         &   0.329  & 0.0203 & 39.51        \\  
    DiffusioNeRF   &   19.94            &   0.683              &   0.296                 & 0.0198             &  50.03  \\ 
    *SinNeRF       &   16.86                &   0.373                  &   0.558                     & 0.0458                  &  31.54  \\ 
    *SPIn-NeRF     &   20.40            &   0.672              &   0.284                 & 0.0156             &  51.84  \\     
    *SDS           &   20.56                &   0.654                  &   0.338                     & 0.0351                  &  49.35  \\ 
    Ours           &   \textbf{20.76}            &   \textbf{0.688}              &   \textbf{0.269}                 & \textbf{0.0154}             &  \textbf{54.13}  \\ 
\bottomrule
\end{tabular}
\end{table}
\begin{table}[t]

\centering
\caption{Quantitative comparison of few-shot view synthesis~\cite{niemeyer2022regnerf}.}%

\label{tab:quantitative_llff_original_protocol_rebutt}
\centering
\small
\setlength{\tabcolsep}{1pt}
\begin{tabular}{l|cccc}
\toprule
    Metrics        &   ~DiffusioNeRF~ &   ~Sparf~ &   ~FreeNeRF~ &  ~Ours~\\ 
\midrule
    PSNR $\uparrow$ &   19.79           &   20.20         &   19.63  & \textbf{21.17}        \\ 
    SSIM $\uparrow$ &    0.568    & 0.630   & 0.612 & \textbf{0.719}\\
    LPIPS $\downarrow$&  0.338           &   0.383         &   0.308  & \textbf{0.264}        \\ 
\bottomrule
\end{tabular}
\end{table}
\begin{table}[t]
\caption{Ablation study.}%
\label{tab:ablations}
\centering
\small
\setlength{\tabcolsep}{2pt}
\begin{tabular}{l|ccc}
\toprule
    Methods &      LPIPS $\downarrow$    &   KID $\downarrow$ &   MUSIQ $\uparrow$    \\ 
\midrule
    BaseNeRF                       & 0.323 & 0.0220  &  49.31 \\ 

    w/ pretrained $\Psi^{\text{inpaint}}$  & 0.291 & 0.0158  &  52.90 \\ 
    w/ fine-tuned $\Psi^{\text{inpaint}}$  & 0.282 & 0.0155  &  53.15 \\ 
    w/ fine-tuned $\Psi^{\text{inpaint}}$ \& $\Psi^{\text{enhance}}$  & \textbf{0.269} & \textbf{0.0154}  &  \textbf{54.13} \\ 
\bottomrule
\end{tabular}
\end{table}


\begin{figure*}[h] 

\captionsetup[subfigure]{labelformat=empty}
\centering
\begin{subfigure}{.245\textwidth}
  \centering
  \includegraphics[width=\linewidth]{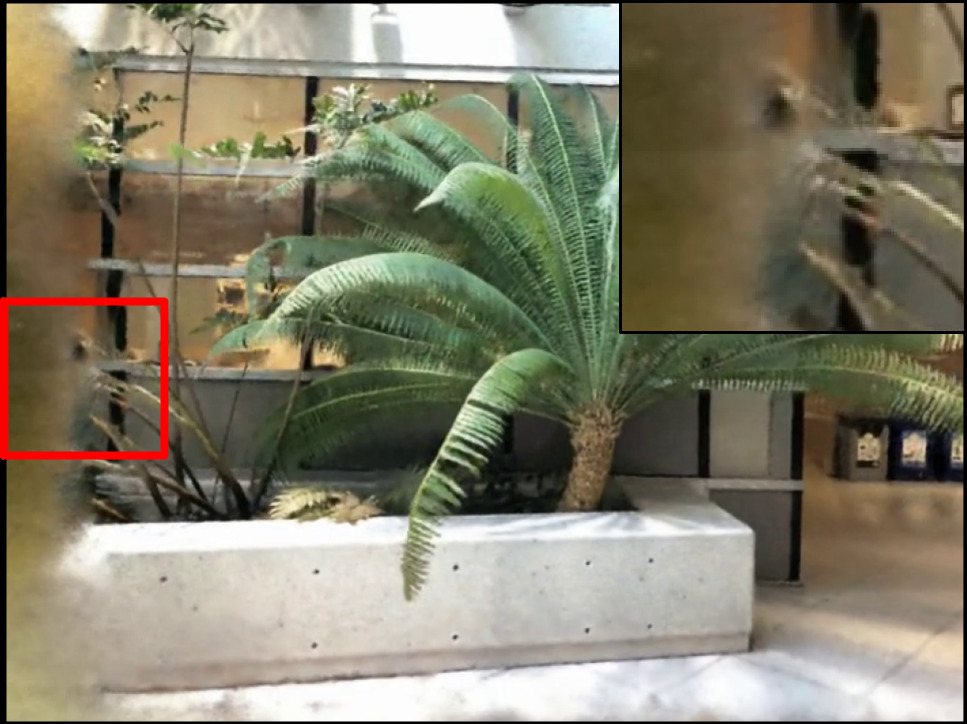}
\end{subfigure}\hfill%
\begin{subfigure}{.245\textwidth}
  \centering
  \includegraphics[width=\linewidth]{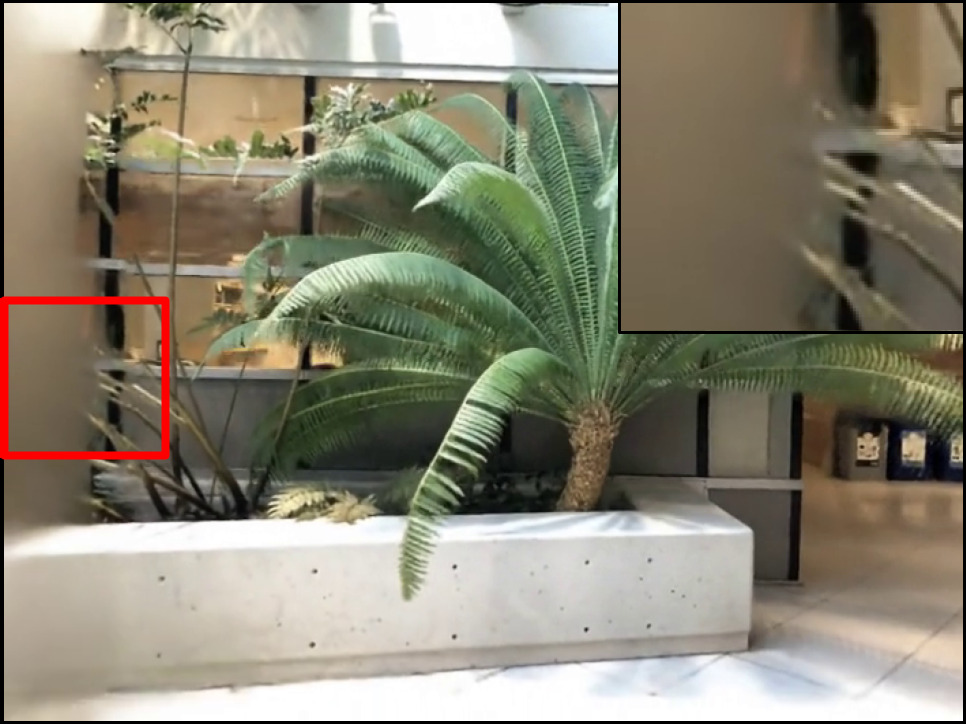}
\end{subfigure}\hfill%
\begin{subfigure}{.245\textwidth}
  \centering
  \includegraphics[width=\linewidth]{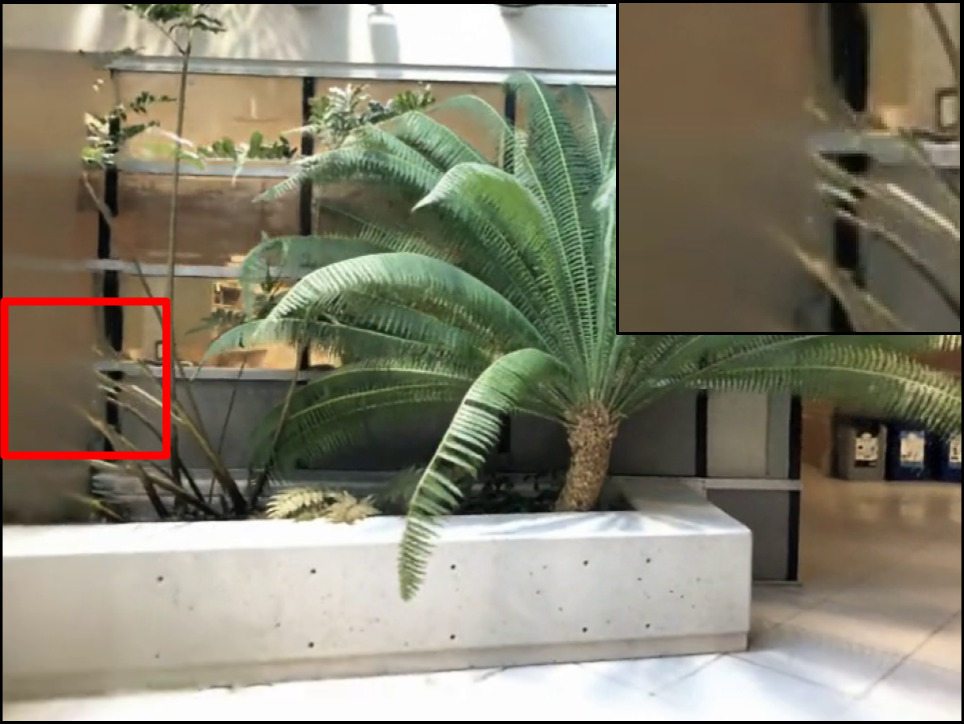}
\end{subfigure}\hfill%
\begin{subfigure}{.245\textwidth}
  \centering
  \includegraphics[width=\linewidth]{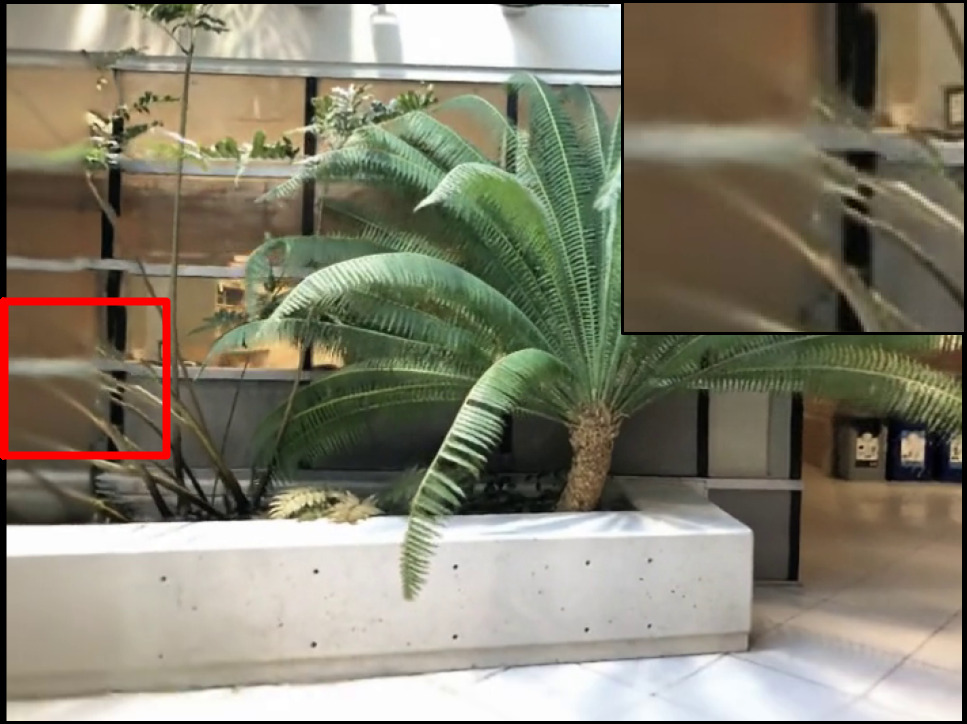}
\end{subfigure}\hfill
\begin{subfigure}{.245\textwidth}
  \centering
  \includegraphics[width=\linewidth]{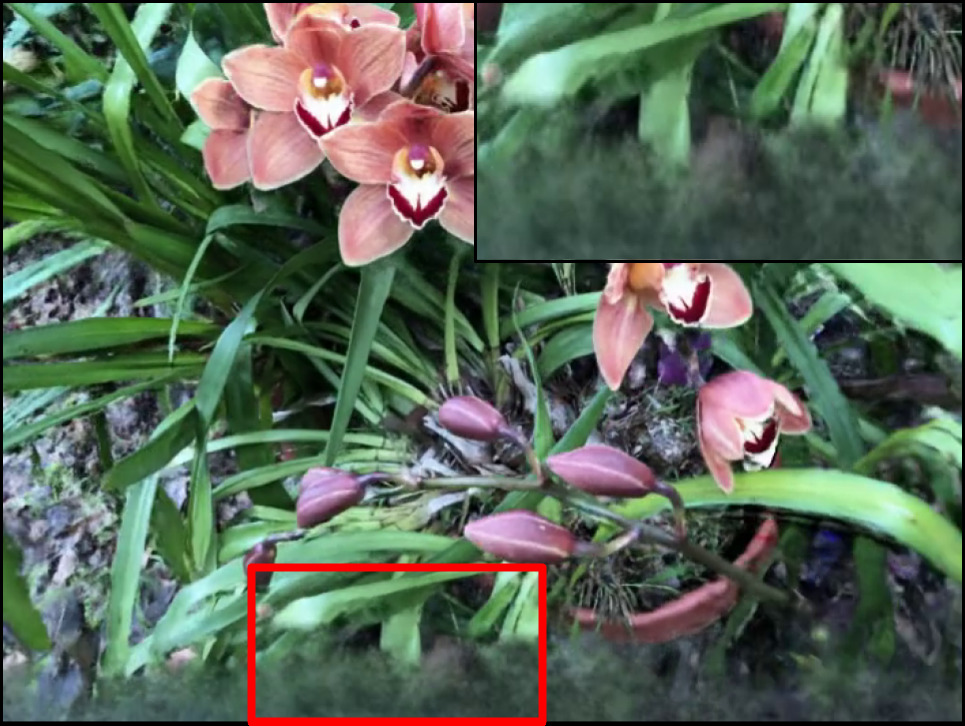}
\caption{\phantom{$\Psi^{\text{inpaint}}$}BaseNeRF\phantom{$\Psi^{\text{inpaint}}$}}
\end{subfigure}\hfill%
\begin{subfigure}{.245\textwidth}
  \centering
  \includegraphics[width=\linewidth]{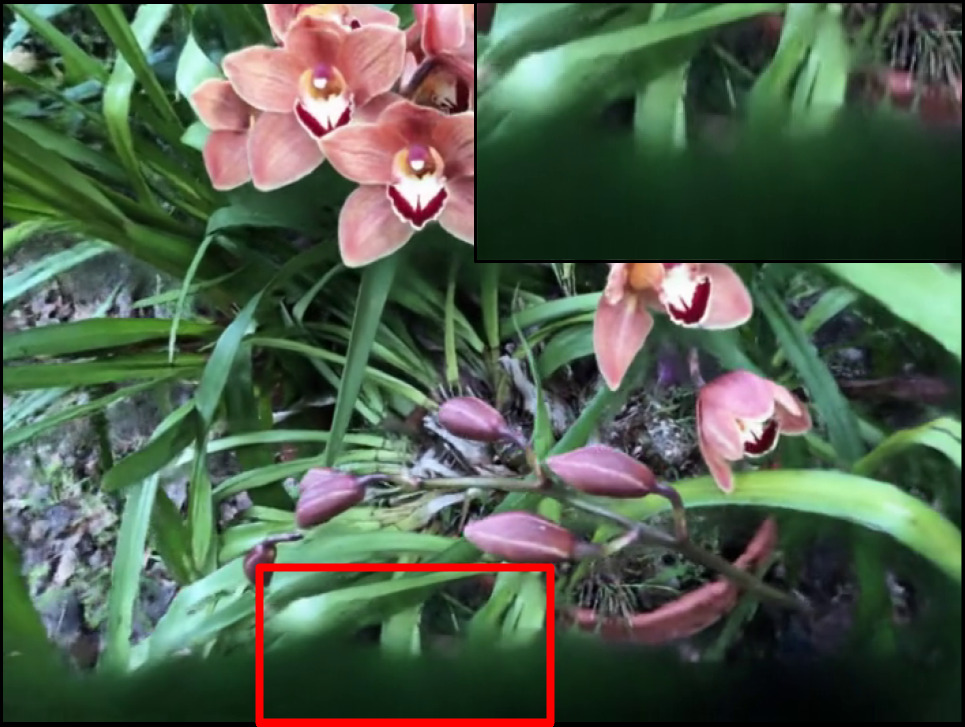}
  \caption{w/ pretrained $\Psi^{\text{inpaint}}$}
\end{subfigure}\hfill%
\begin{subfigure}{.245\textwidth}
  \centering
  \includegraphics[width=\linewidth]{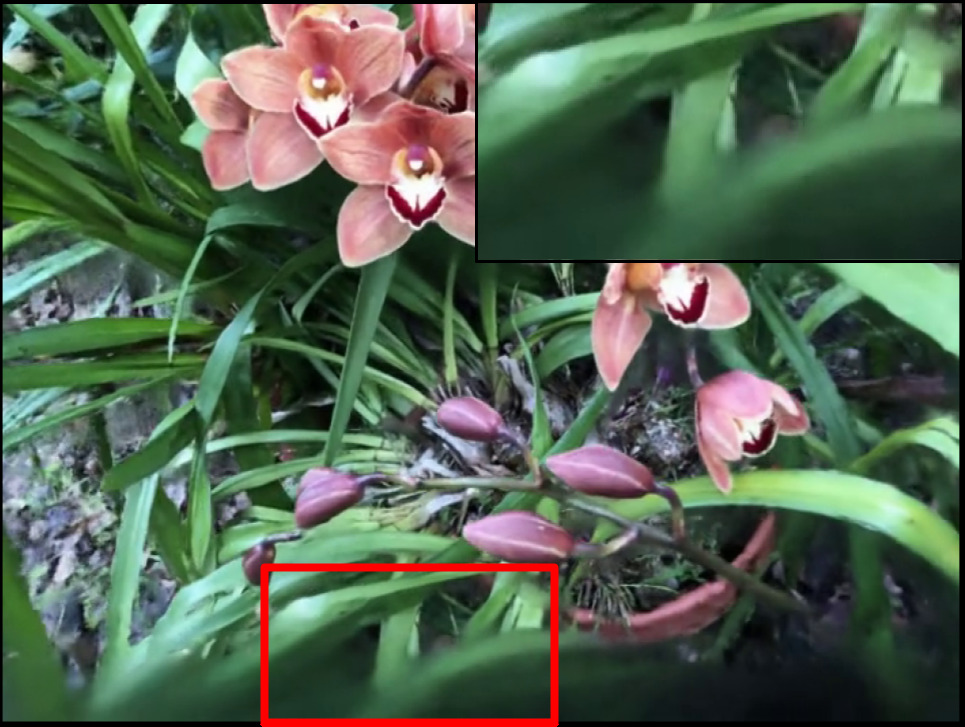}
  \caption{w/ finetuned $\Psi^{\text{inpaint}}$}
\end{subfigure}\hfill%
\begin{subfigure}{.245\textwidth}
  \centering
  \includegraphics[width=\linewidth]{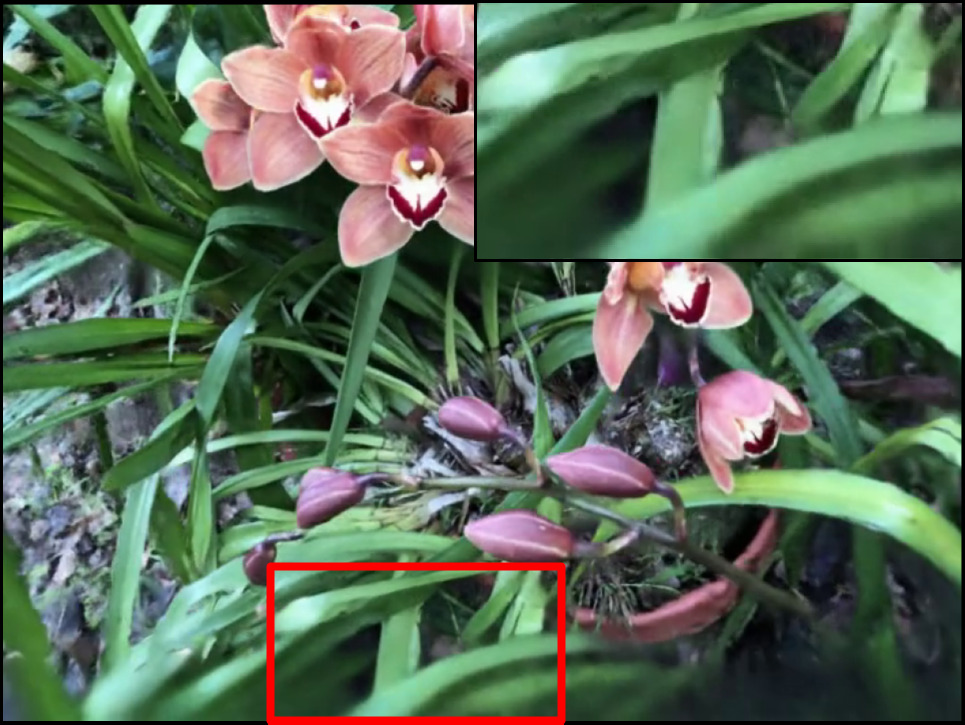}
  \caption{w/ finetuned $\Psi^{\text{inpaint}}$\&$\Psi^{\text{enhance}}$}
\end{subfigure}\hfill
\caption{\textbf{Ablation study.}
Inpainting disoccluded regions with a pretrained diffusion model results in blurry and color-drifted outcomes. However, fine-tuning the model on specific scenes reduces these issues, as the fine-tuned model captures scene-specific statistics more accurately. Additionally, our enhancement model further enables fine-grained details in NeRF-rendered images, producing sharper results.
}
  \label{fig:ablations}
\end{figure*}
\begin{figure*}[htbp] 
  \centering
\includegraphics[width=\linewidth]{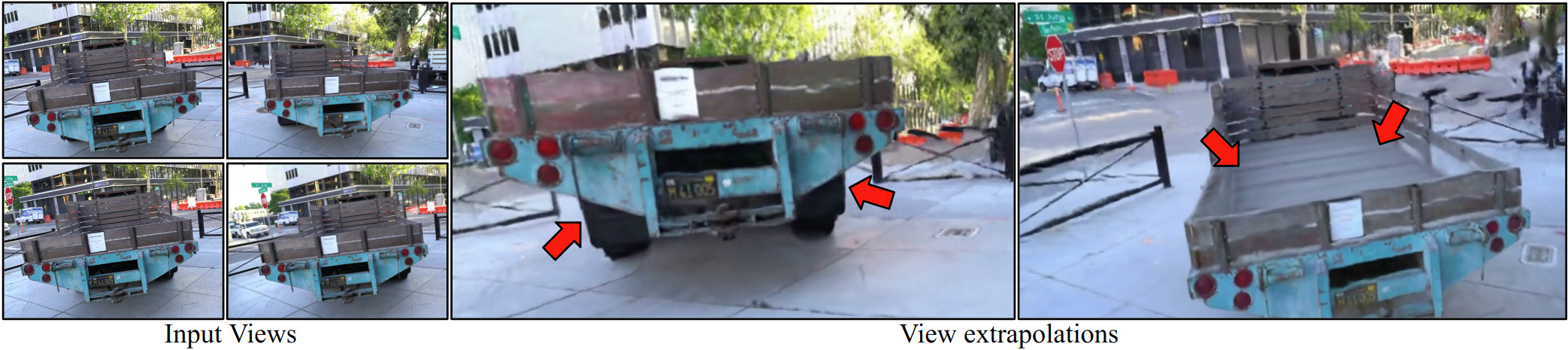}
\caption{
\textbf{Qualitative results on Tanks\&Temples.} 
While only a very small portion of the deck and the tires of the truck is visible from input viewpoints (left), our model is still able to synthesis the missing content (right).}
\label{fig:qual_tNt}
\end{figure*}
In Table~\ref{tab:quantitative_llff}, our method surpasses related works across various metrics, showcasing our approach's superior ability to inpaint unseen regions in view extrapolation tasks. Furthermore, Figure~\ref{fig:qualitative} presents a qualitative comparison, highlighting the distinctions between our method and competing approaches.

While Sparf and FreeNeRF demonstrate proficiency in estimating geometry and appearance for regions captured by input viewpoints, they fall short in generating meaningful content for view extrapolation scenarios. DiffusioNeRF, sharing our utilization of a diffusion prior to enhance NeRF quality, is limited by its patch-based model's narrow receptive field, preventing the synthesis of coherent content. Our diffusion model, in contrast, processes the entire image to generate meaningful and consistent content. *SPIn-NeRF, employing perceptual loss to address inconsistencies in supervision from inpainted images, inadvertently introduces pattern artifacts. Additionally, while SDS loss can produce reasonable content, it often lacks the complexity of detail.

Compared to these methods, our technique excels in creating believable content that is both stylistically consistent and detailed.
\paragraph{Comparison of few-shot view synthesis: }
In Table~\ref{tab:quantitative_llff_original_protocol_rebutt}, we demonstrate that our method outperforms other baselines in the few-shot view synthesis protocol with only 3 training views. This indicates that our approach can significantly reduce the number of required training views.
\paragraph{Ablations: }
In Figure~\ref{fig:ablations}, we illustrate the impact of removing components from our pipeline on the task of view extrapolation. The pretrained inpainting model struggles to fill masked regions with content that maintains consistent appearance and structure, leading to results that exhibit blurriness and color drift in NeRF, as depicted in Figure~\ref{fig:ablations}. By fine-tuning the inpainting model with the specific scene's captured photos, the diffusion model learns the scene's unique distribution, enabling it to more accurately generate content with consistent structure and appearance. Moreover, our enhancement model is capable of adding even greater detail than the fine-tuned inpainting model. Additionally, the results in Table~\ref{tab:ablations} further demonstrate the effectiveness of each component within our pipeline.

\subsection{Tanks \& Temples}
\begin{table}[t]
\caption{Quantitative comparison on Tanks\ \&\ Temples Dataset.}%
\label{tab:quantitative_tNt}
\centering
\small
\setlength{\tabcolsep}{2pt}
\begin{tabular}{l|ccccc}
\toprule
    Methods &   PSNR $\uparrow$    &   SSIM $\uparrow$   &   LPIPS $\downarrow$    &   KID $\downarrow$ &   MUSIQ $\uparrow$    \\ 
\midrule
    Sparf     &   \textbf{19.21}  &   0.576  &   0.518 & 0.149  & 36.07 \\ 
    *SPIn-NeRF &   18.28  &   0.683  &   0.348 & \textbf{0.042}  & 37.63  \\ 
    Ours       &   19.20  &   \textbf{0.718}  &   \textbf{0.312} & 0.065  & \textbf{40.85}  \\ 
\bottomrule
\end{tabular}
\vspace{-4.5mm}
\end{table}

Given that this dataset features a significant proportion of pixels with extremely large depth values, we limit our comparison to methods equipped to handle unbounded scenes. In Table~\ref{tab:quantitative_tNt}, our method surpasses others in LPIPS and MUSIQ scores, signifying superior visual quality of our results. However, our KID score falls short of SPIn-NeRF's. We hypothesize that this is likely due to the dataset's small size that may be insufficient to accurately estimate the test set's distribution. Furthermore, we showcase an example highlighting our model's ability to synthesize extensive areas of content that are unobservable from the input viewpoints (see Figure~\ref{fig:qual_tNt}).


\section{Conclusion}
In this paper, we present an approach to broadening viewing range for a NeRF captured from a small, narrowly grouped set of input images. Our method, dubbed as ExtraNeRF, uses a pretrained diffusion model to produce new detail in two ways: first to inpaint given a visibility mask computed from the NeRF itself, then to enhance detail. We find that per-scene fine-tuning, design of enhancement model, and our data collections are critical for achieving good results. We set a new SOTA for view extrapolation on the LLFF dataset and Tanks \& Temples dataset. 

\vspace{-3mm}
\paragraph{Acknowledgment: } This work was supported by the UW Reality Lab, Meta, Google, OPPO, and Amazon.
\vspace{-8.5mm}
{
    \small
    \bibliographystyle{ieeenat_fullname}
    \bibliography{main}
}

\end{document}